%% file: main.tex
\PassOptionsToPackage{numbers,sort}{natbib}
\documentclass[final,5p,twocolumn]{elsarticle}  

\input{packages}

\input{commands}

\journal{Pattern Recognition}

\begin{document}
\date{November 5, 2022}  

\begin{frontmatter}

\title{\vspace*{-15mm}Real-Time Siamese Multiple Object \\ Tracker with Enhanced Proposals\vspace*{-2mm}}

\author[mymainaddress]{Lorenzo~Vaquero\corref{mycorrespondingauthor}}
\cortext[mycorrespondingauthor]{Corresponding author}
\ead{lorenzo.vaquero.otal@usc.es}

\author[mymainaddress]{V\'{i}ctor~M.~Brea}
\ead{victor.brea@usc.es}

\author[mymainaddress]{Manuel~Mucientes}
\ead{manuel.mucientes@usc.es}

\address[mymainaddress]{Centro Singular de Investigaci\'{o}n en Tecnolox\'{i}as Intelixentes (CiTIUS), \\ Universidade de Santiago de Compostela, Santiago de Compostela, Spain\vspace*{-5mm}}

\input{abstract}

\end{frontmatter}

\input{introduction}

\input{related-work}

\input{net-architecture}

\input{experimentation}

\input{conclusions}

\input{thanks}

\bibliography{bib}

\input{biography}

\end{document}

%% file: packages.tex
\usepackage[T1]{fontenc}  
\usepackage[cmintegrals]{newtxmath}
\usepackage{bm}  

\usepackage{xspace} 
\usepackage[table,xcdraw]{xcolor}
\usepackage{hhline}
\usepackage{makecell}
\usepackage{picins}
\usepackage{chngpage}
\usepackage{multirow} 
\usepackage{wasysym}

\graphicspath{{figures/}}

\usepackage[inline]{enumitem}
\usepackage[ruled,norelsize]{algorithm2e} 

\makeatletter
\newcommand{\removecolumnlatexerror}{\let\@latex@error\@gobble}
\makeatother

\usepackage{array}

\usepackage[caption=false,font=footnotesize]{subfig}

\usepackage{url}
\usepackage{hyperref}
\hypersetup{pdfauthor="Lorenzo Vaquero"}

%% file: commands.tex
\newcommand{\siammt}{SiamMOTION\xspace}

\newcommand{\new}[1]{\ensuremath{\tilde{#1}}}  
\newcommand{\N}{\ensuremath{\mathcal{N}}\xspace} 
\newcommand{\Scales}{\ensuremath{\mathcal{S}}\xspace}

\newcommand{\xcorr}{\ensuremath{\APLstar}\xspace}  
\newcommand{\crop}{\ensuremath{\kappa}\xspace}  

\newcommand{\xcorrnew}{\ensuremath{\new{\xcorr}}\xspace}

\newcommand{\bigO}{\mathcal{O}}

\newcommand\mypound{\scalebox{0.8}{\raisebox{0.4ex}{\fontseries{b}\selectfont\#}}}

\definecolor{exemplargreen}{HTML}{f3d6d7}
\definecolor{searchred}{HTML}{f3d6d7}
\definecolor{darkgreen}{HTML}{445b00}

%% file: abstract.tex

\begin{abstract}
Maintaining the identity of multiple objects in real-time video is a challenging task, as it is not always feasible to run a detector on every frame.
Thus, motion estimation systems are often employed, which either do not scale well with the number of targets or produce features with limited semantic information.
To solve the aforementioned problems and allow the tracking of dozens of arbitrary objects in real-time, we propose SiamMOTION.
SiamMOTION includes a novel proposal engine that produces quality features through an attention mechanism and a region-of-interest extractor fed by an inertia module and powered by a feature pyramid network.
Finally, the extracted tensors enter a comparison head that efficiently matches pairs of exemplars and search areas, generating quality predictions via a pairwise depthwise region proposal network and a multi-object penalization module.
SiamMOTION has been validated on five public benchmarks, achieving leading performance against current state-of-the-art trackers. Code available at: \url{https://github.com/lorenzovaquero/SiamMOTION}
\end{abstract}


%% file: introduction.tex

\section{Introduction\label{sec:introduction}}

Visual object tracking consists in maintaining the identity of one or more targets throughout a video.
This is among the first steps in video analytics applications, enabling systems to carry out functions that range from video surveillance to robot navigation~\cite{Fang2021}.
Traditionally, multiple object tracking (MOT) has been addressed through the association of detections.
Namely, for each new frame, a pre-trained detector is run in order to locate all the objects of interest in the scene.
Following this, the current detections are associated with those of the previous frame, revealing the displacements and scale changes of the objects of interest.

However, when considering the use of a tracker to tackle a real-world problem there can be numerous constraints to consider.
One of the most common and, at the same time, one of the most challenging restrictions is the real-time processing of the video, as traditional MOT systems are computationally very expensive.
Amongst all the methods proposed at the MOT2020 Challenge~\cite{Dendorfer2020}, only 5 are able to run in real-time without considering the detector time.
If we also take into account the detector runtimes, all these approaches should be discarded\footnote{We consider that a system/module operates in real-time if it runs at least at 25~fps for HD720 resolution on an NVIDIA TITAN V or similar.}, as accurate detectors alone already struggle to keep up with real-time ---EfficientDet-D3~\cite{Tan2020} runs at $23$~fps on an NVIDIA TITAN~V for HD720 images.

\begin{figure}[!t]
\centering
\includegraphics[width=0.85\linewidth]{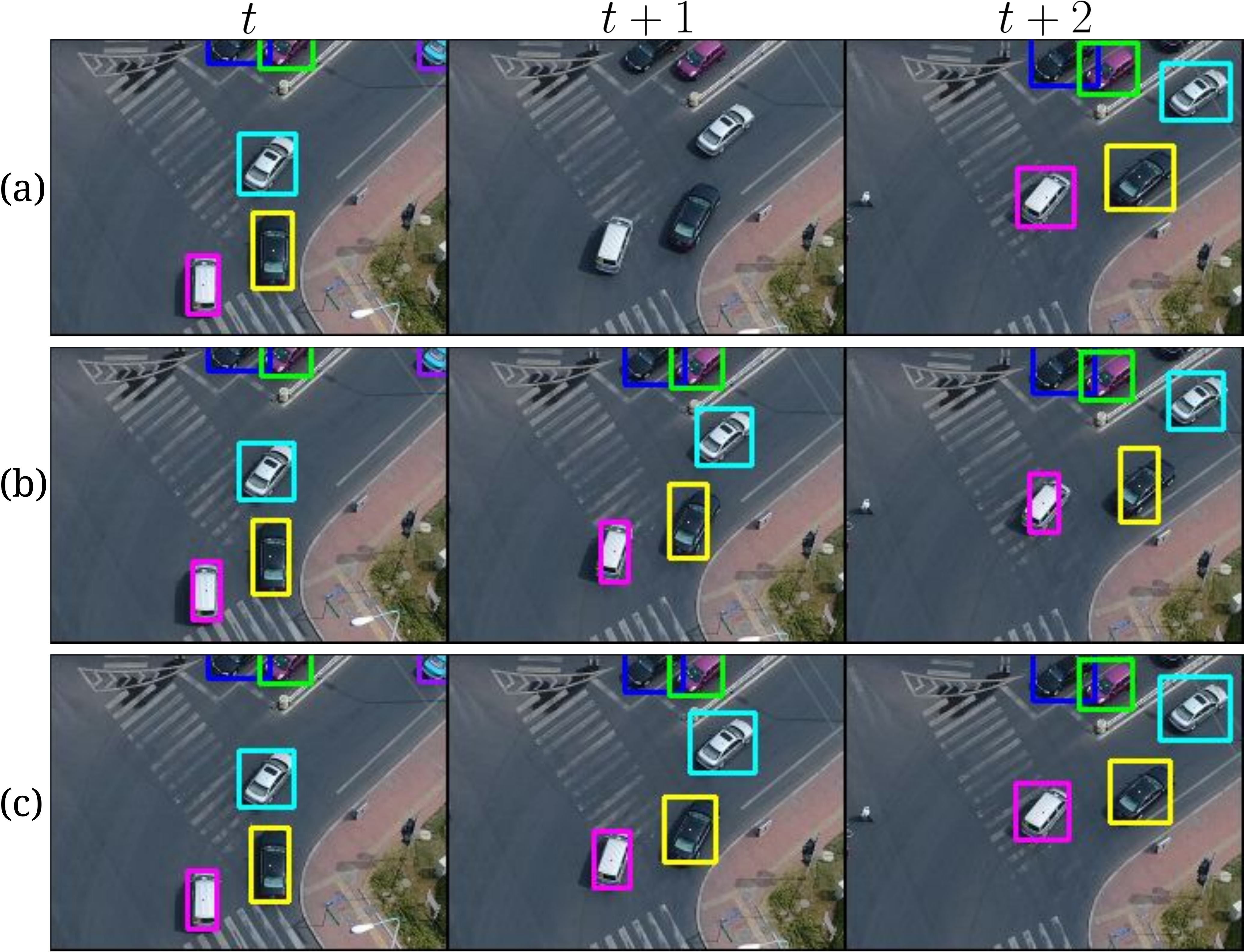}
\vspace*{-2mm}
\caption{Comparison of (a)~a multi-object tracking system lacking motion estimation between detections, (b)~SiamMT~\cite{Vaquero2021}, and (c)~our proposal (\siammt).
Notice how (a) is unable to handle all frames and (b) does not detect changes in the aspect ratio of the objects.\vspace*{-5mm}}
\label{fig:qualitative}
\end{figure}

Thus, it is necessary to adopt mechanisms capable of providing the position of the objects in all the frames without relying on continuous detections (\figurename~\ref{fig:qualitative}).
This is what is often referred to as \textit{motion estimation} between detections, which nowadays is powered by Visual Object Tracking (VOT) approaches, giving rise to the concept of multiple visual object tracking (MVOT).
These methods are applied across various types of systems, as they benefit from the latest advances in single-object tracking~\cite{Ciaparrone2020}.
However, since these trackers are designed for a single target and their multiple instantiation is costly, such solutions are only suitable for uncrowded scenarios.
To tackle this issue, the approach of addressing the problem more globally arises, trying to share as many computations as possible between objects~\cite{Vaquero2021}.
This allows to keep up with several dozens of targets in real-time while applying single-object tracking approaches.
Still, the full potential of this concept has not yet been exploited.

In order to expand the current trend of visual object trackers for motion estimation we propose \siammt (\textbf{Siam}ese \textbf{M}ultiple \textbf{O}bject \textbf{T}racker with \textbf{I}nertia and attenti\textbf{O}n \textbf{N}etwork).
\siammt relies on the fundamentals defined in~\cite{Vaquero2021} for tracking multiple objects in an efficient and scalable manner and integrates them with the latest single object tracking methods, all while solving some core problems of the aforementioned architecture.
\siammt's architecture includes a proposal engine~(PE) that integrates an inertia module, a region-of-interest extractor, and an attention mechanism; and a comparison head~(CH) composed of a region proposal network and a multi-object penalization module.
The main novelties of our proposal are summarized as follows:

\begin{itemize}
  \item The feature extraction is performed through a feature pyramid network (FPN) that allows to obtain meaningful features for all object sizes.
  Search areas are extracted from these features using an inertia module that takes into account the previous positions of each object.
  To the best of our knowledge, this is the first time these methods are employed in a detection-independent visual object tracker.
  
  \item A lightweight attention mechanism is employed to enhance the features most closely related to the objects of interest.
  This, together with a pairwise-depthwise region proposal network (PD-RPN), allows the prediction of accurate bounding-boxes in real-time.
  
  \item A novel multi-object penalization module is used to suppress distractors and outliers by taking into account all objects of interest in the scene.
  It models interactions between targets and applies 4 different types of penalizations, each one devoted to addressing a specific type of tracking error.
  
  \item We validate our proposal on five public datasets using VOT-RT metrics~\cite{Kristan2019}, achieving leading performance against current state-of-the-art trackers.
\end{itemize}

%% file: related-work.tex

\section{Related work\label{sec:related-work}}

\subsection{Motion estimation}
Traditional multi-object tracking systems depend heavily on the quality of the detector employed, requiring accurate ---and therefore costly--- detectors to reach their full potential~\cite{Fernandez-Sanjurjo2021}.
This poses a problem when there are real-time or hardware constraints, as it is not feasible to obtain detections for every frame.
In this paper, we propose a novel multiple visual object tracker (MVOT) architecture to address this limitation.
Such models can be integrated into full tracking systems, being able to estimate the motion of objects for those frames with no detections.

Initially, the preferred approach for estimating the motion of multiple targets between detections was through Bayesian filters~\cite{Bewley2016}.
However, current MVOT methods nowadays perform motion estimation through visual single-object tracking methods~\cite{Ciaparrone2020}.
The most straightforward approach consists in instantiating a new individual tracker for each object that appeared in the scene.
This can be carried out either by including the tracker as an independent component~\cite{Zhou2020a} or by embedding it into the architecture~\cite{Yin2020}.
However, the caveat is that these methods cause the system to slow down with each additional object, so they are only suitable when the number of targets is small.

In order to allow the tracking of several dozens of objects and alleviate the problems mentioned above, \cite{Vaquero2021} emerges.
This tracker aims to share the most expensive operations of the architecture between the objects, reusing feature computations.
This, combined with the new operators it introduces, results in most of the network having a constant computational cost, regardless of the number of objects.
Thus, while~\cite{Yin2020} runs at $5$~fps for an FHD video with $21$~objects, \cite{Vaquero2021} is capable of handling $100$~objects at $25$~fps.
However, these approaches still present some problems and do not currently benefit from the latest advances in single-object tracking.

\begin{figure*}[!t]
\centering
\includegraphics[width=0.95\linewidth]{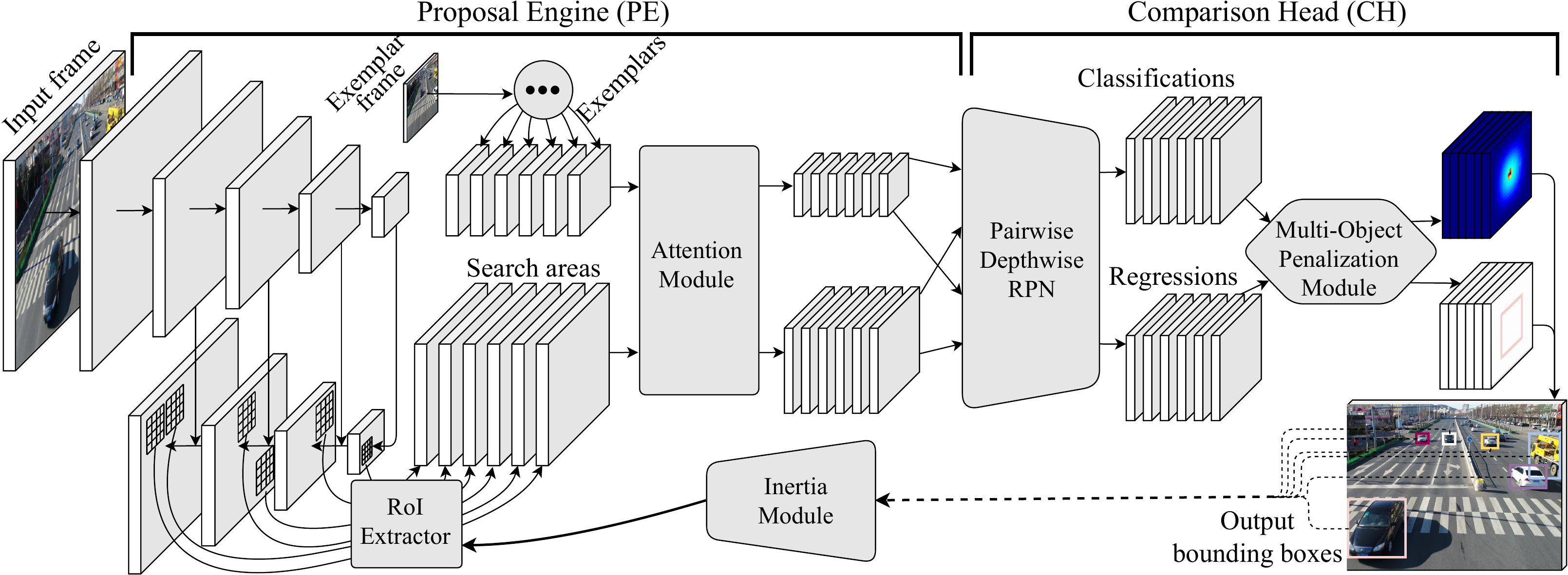}
\vspace*{-1mm}
\caption{\siammt's architecture.
First, it extracts the global features of the input frame.
Then, it obtains the search area features of each object through the RoI~Extractor, using the locations provided by the Inertia Module.
The exemplar features are computed analogously at the beginning of the sequence and are reused throughout the rest of the video.
Next, the attention mechanism is applied over these tensors and the previously extracted exemplars, before comparing them using the Pairwise-Depthwise RPN.
Finally, the RPN output is fed through the Multi-Object Penalization Module to obtain refined bounding boxes for each object.
}
\label{fig:architecture}
\end{figure*}

\subsection{Visual Object Tracking}
Motion estimation between detections is mainly performed using single-object online trackers.
They comprise an initialization step, in which they receive the exemplar appearance of the object in order to integrate it into a similarity function.
Then, for each new frame, this similarity function looks for the object of interest, reporting its new bounding box.
Originally, this task was carried out using Discriminative Correlation Filters (DCF) which, by representing the object with a single filter, were able to distinguish the background from the target~\cite{Bolme2010}.
These approaches became increasingly sophisticated, modeling these filters as convolutional layers~\cite{Yuan2020} and applying optimization frameworks to speed up the learning process~\cite{Xu2020a}.

However, the state of the art in tracking is currently driven by deep learning approaches.
These trackers implement their similarity function through deep convolutional neural networks.
Given the need to compare an exemplar image with a search area defined in the frame, Siamese neural networks emerge as the most natural choice, with~\cite{Bertinetto2016} being the precursor of the current state of the art.
To further enhance this architecture, contributions from other fields of computer vision were gradually incorporated.
Thus, there are trackers allowing changes in the aspect ratio of the bounding box ---using a Region Proposal Network (RPN)~\cite{Li2018} or anchor-free~\cite{Guo2020}---, employing more sophisticated backbones~\cite{Li2019}, featuring attention modules~\cite{Yu2020b}, with segmentation information~\cite{Yin2021}, or including branches for estimating the proposal quality~\cite{Xu2020}.

Within the CNN approaches, some trackers somewhat deviate from the traditional Siamese formula of~\cite{Bertinetto2016}.
For example, \cite{Danelljan2019} presents an architecture with dedicated components for object classification and estimation.
The classification module specializes in discriminating between distractors, while the estimator component tries to predict the overlap between the prediction and the object.
These types of networks are nondeterministic and thus are strongly affected by the initialization of the exemplar.
This is why there have been efforts to incorporate modules capable of predicting the quality of such initializations~\cite{Bhat2019}.
The accuracy they offer is good, but as they iteratively refine the bounding box, these methods are computationally expensive and very sensitive to hyperparameters, which discourages their use as MVOT solutions.

%% file: net-architecture.tex

\section{\siammt Network Architecture\label{sec:net-architecture}}
As shown in \figurename~\ref{fig:architecture} \siammt's architecture comprises: a backbone for feature extraction, a proposal engine~(PE) that embeds a Feature Pyramid Network-based region-of-interest extractor, an inertia module for the definition of search areas, and an attention module; and a comparison head~(CH) consisting of a region proposal subnetwork for classification and regression, and a multi-object penalization module.
All these components are addressed in the following subsections.

\subsection{Feature Extractor}
\begin{figure}[!t]
\centering
  \subfloat[]{\includegraphics[width=0.8\linewidth]{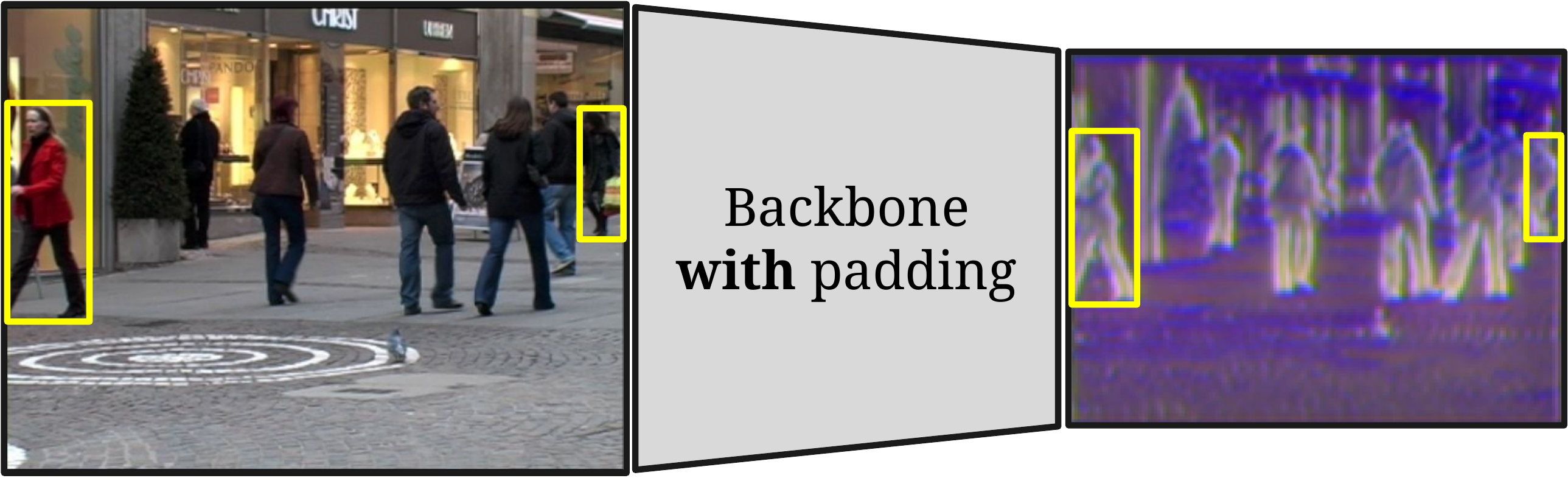}
  \label{fig:padding-resnet}}
\hfil
  \subfloat[]{\includegraphics[width=0.8\linewidth]{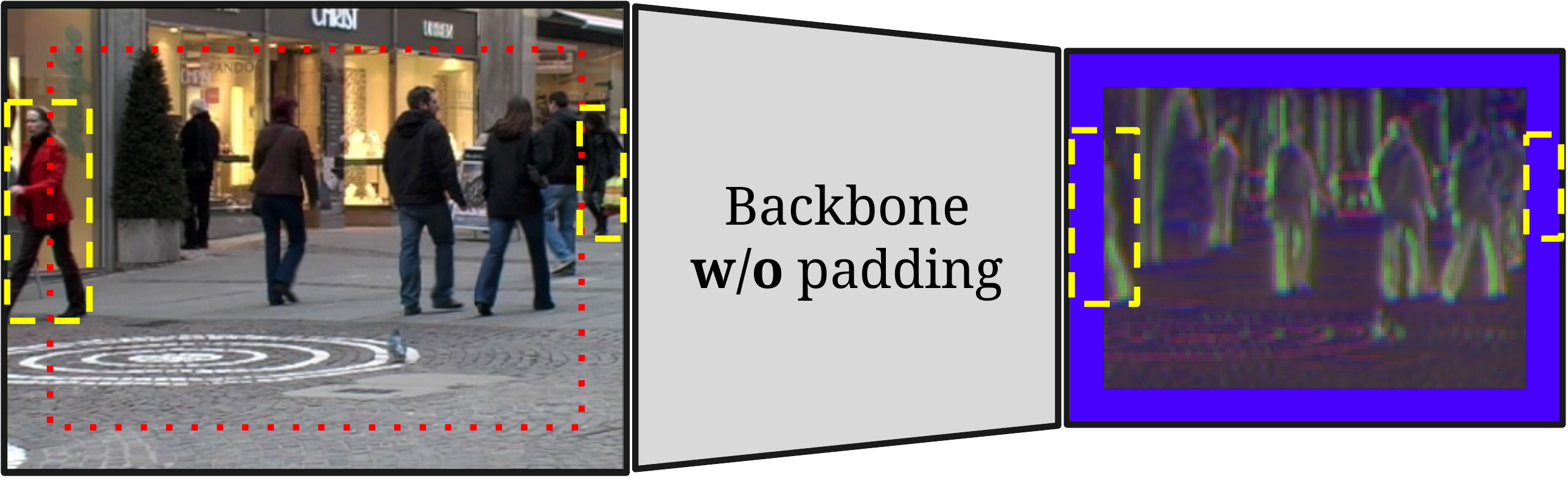}
  \label{fig:padding-alexnet}}
\caption{Features extracted with \protect\subref{fig:padding-resnet}~a backbone with padding and \protect\subref{fig:padding-alexnet}~a backbone without it.
The absence of padding results in a blind area ---outside the red-dotted region--- at the edges of the frame, which produces low-quality features for the highlighted objects.}
\label{fig:padding}
\end{figure}

Feature extractors transform an image from an (usually) RGB color space to a semantic embedding space, in which the various channels encode more meaningful information about the objects in the scene.
Backbones that lack padding dilute the information at the edges of the image (\figurename~\ref{fig:padding}), but this does not pose a problem for single-object trackers since they define their exemplar and search area images with some margin.
However, as \siammt extracts the features of the whole frame, relying on a backbone without padding would result in poorer tracking quality for the targets close to the boundaries.
This is a major problem considering that, in multi-object scenarios, targets enter and exit the scene mostly through the edges of the image.
For this reason, \siammt adopts a padded backbone based on ResNet~\cite{He2016}.
Yet, unlike other Siamese trackers that employ deep padded backbones~\cite{Li2019,Guo2020}, we keep the last convolutional blocks with their standard stride of 2.

\subsection{Proposal Engine}
Visual object trackers generate feature tensors for each search area, which will eventually be compared with the exemplars to locate each object in the scene.
Approaches such as \mbox{SiamRPN\texttt{+}\texttt{+}}~\cite{Li2019} or \mbox{SiamAttn}~\cite{Yu2020b} clip fragments of the input frame and then extract their features, which produces accurate results but deems very slow for several objects.
\mbox{SiamMT}~\cite{Vaquero2021} generates the search areas by cropping the frame features with a modified RoI Align~\cite{He2017}, which improves the efficiency but produces low-quality tensors.
The proposal engine (PE) presented in this paper efficiently generates high-quality features that properly fit the objects and capture their details with adequate resolution.
It integrates an inertia module, a region-of-interest (RoI) extractor, and an attention mechanism.

\subsubsection{Region-of-Interest Extractor}
As previously stated, single-object trackers define a search area in the frame.
This area is cropped and resized to a fixed size ---usually $255 \ \text{px}^2$ or $271 \ \text{px}^2$--- before feeding it to the neural network.
As a result, the target is displayed with a nearly constant size, which is the reason why~\cite{Li2018} and its evolutions do not need to take into account different anchor sizes.
However, since \siammt extracts the features of the whole frame for several simultaneous objects, it cannot take advantage of this property.
In~\cite{Vaquero2021} a new operator is proposed for the creation of search areas in the aforementioned conditions, yet it does not perform well with very small or very large objects, sometimes making it necessary to rescale the entire frame.

In order to obtain meaningful features for all search area sizes, we implement the region-of-interest (RoI) crop-and-resize operation through a feature pyramid network (FPN).
Thus, instead of relying just on the final layer of the backbone, some of its intermediate layers also serve as inputs, capturing the scene at different resolutions.
However, as higher-resolution layers contain poorer semantic information, we integrate them into a feature pyramid network to produce rich multi-scale representations.
To the best of our knowledge, this is the first time such an approach is employed in a detection-independent object tracker.

Our RoI extraction mechanism has several similarities with those found in object detectors~\cite{Lin2017}.
However, it presents a number of important differences:
\begin{itemize}
    \item The input coordinates for the RoI extractor are not provided by a proposal generator based on the frame features, but by an inertia module that predicts the positions of objects based on their motion (Section~\ref{sec:inertia-module}).
    
    \item The regions covered by the RoI extractor are always square.
    This ensures that the objects are always depicted with the same aspect ratio relative to the frame (1:1), which improves the learning of the similarity function.
    
    \item The choice of the pyramid level $k$ from which to extract the features is calculated as:
    \begin{equation} \label{eq:fpn_level_selection}
    k = \left\lfloor k_0 + \log_2\left(\frac{\sqrt{A_{w}A_{h}}}{8 \lfloor 255/8\rfloor} \right) \right\rfloor
    \end{equation}
    Here $255$ and $8$ are the canonical sizes of the search area and network stride in SiamFC-based trackers, respectively, $k_0$ is the FPN level with a resolution of $1/8$ (i.e., level $3$), and $A_{w}$ and $A_{h}$ are the area's dimensions.
    In our architecture we consider values of $k$ in the range [1, 4].
    
    \item The cropping and resizing of the region is performed using RoI Align~\cite{He2017} with one sampling point per bin, for better computational efficiency.
    Following this operation, a $1\times1$ convolution is applied to transform the maps to a comparable representation, regardless of their original resolution.
\end{itemize}

The  proposed RoI extractor yields tensors that will always depict the targets with an approximately constant size.
This, along with the fact that the FPN provides a good feature hierarchy, means that tensors generated at different levels will have rich semantic information and will be comparable with each other.
Therefore, it allows the rest of the network architecture to be independent of the level $k$ from which each search area has been extracted ---as object detectors that rely on a single shared class/box prediction head for all resolution levels do---, requiring fewer learned parameters and simplifying the convergence of the algorithm.
This final remark is particularly relevant, as it makes our approach the first object tracker that compares features extracted at different resolutions, since other systems that apply multi-layer similarities ---e.g., \mbox{SiamCAR}~\cite{Guo2020} or \mbox{SiamRPN\texttt{+}\texttt{+}}~\cite{Li2019}--- adapt their backbones so that the considered layers have the same spatial resolution.

\subsubsection{Inertia Module\label{sec:inertia-module}}
For every new frame, single object trackers define an area to search for the object.
The reason is twofold:
\begin{enumerate*}[label=(\roman*)]
  \item to keep the object constant in size ---we address this with more detail in the next section---;
  \item and to avoid scanning the entire frame, as this would be computationally very expensive.
\end{enumerate*}
Since its introduction in~\cite{Bertinetto2016}, following networks have adopted the same na\"ive approach for the definition of this area, centering it on the last known position of the object.
This method provides good results but presents problems when the target moves fast or is very small ---\cite{Li2018} alleviates it by making the search area larger when these cases show up, but carries a higher cost.
To solve this problem, we propose the use of an inertia module that smartly adjusts the placement of the search area.

\begin{figure}[!t]
\centering
\includegraphics[width=0.80\linewidth]{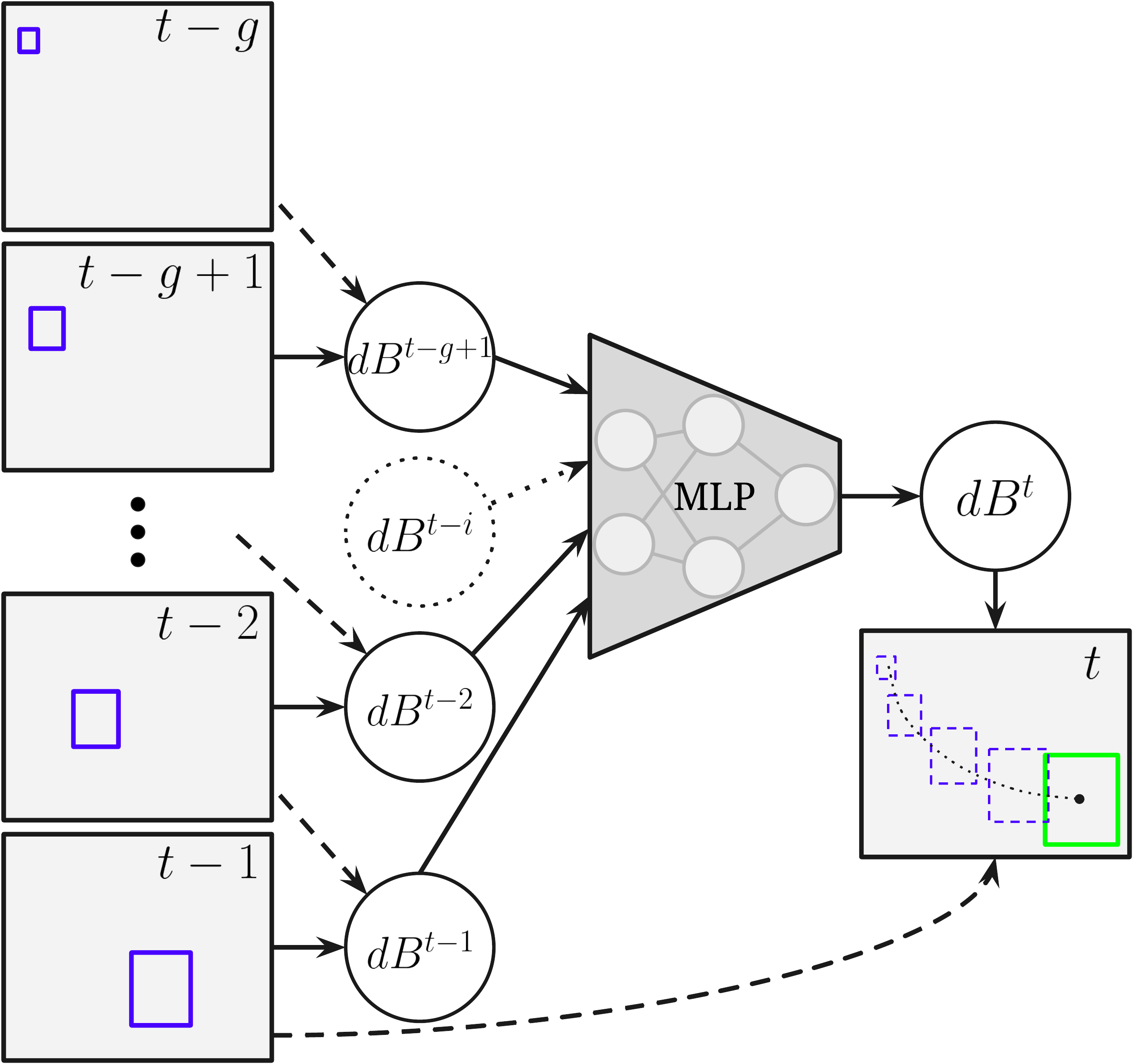}
\vspace*{-2mm}
\caption{\siammt's inertia module.
Using a multilayer perceptron (MLP), it is able to make a coarse prediction of the object's future position based only on its previous coordinates.\vspace*{-1mm}}
\label{fig:mlp}
\end{figure}

The inertia module makes a coarse prediction of the object's location and size at timestamp $t$ ---i.e., $B^t = \{(B_{x}^{t}, B_{y}^{t}, B_{w}^{t}, B_{h}^{t})\}$--- solely from its coordinates in the previous frames, without resorting to visual information.
Thus, it is fed by the predictions of the multi-object penalization module to roughly estimate the position of the object in the future.
The inertia module is implemented through a multilayer perceptron (MLP) that receives the differences between consecutive coordinates $dB$ for the last $g$ video frames (\figurename~\ref{fig:mlp}).
These differences between pairs of coordinates are computed as bouding-box regressions, considering the previous instant as the anchor:
\begin{equation}
\begin{aligned}
dB_{x}^{t-i} &= \frac{B_{x}^{t-i} - B_{x}^{t-i-1}}{B_{w}^{t-i-1}}, \qquad & dB_{y}^{t-i} = \frac{B_{y}^{t-i} - B_{y}^{t-i-1}}{B_{h}^{t-i-1}} \\
dB_{w}^{t-i} &= \log_e\left({\frac{B_{w}^{t-i}}{B_{w}^{t-i-1}}}\right),  & dB_{h}^{t-i} = \log_e\left({\frac{B_{h}^{t-i}}{B_{h}^{t-i-1}}}\right)
\end{aligned}
\end{equation}
where $i \in [1, g)$.
These parameterizations ensure scale-invariant locations and positive-sized bounding boxes.

Using a neural network provides an advantage over other classical models, as it is able to deal with noise in the input and easily adapt to the tracker's particular behavior.
Moreover, as it is fast and lightweight, its predictions can be used as a fallback mechanism if the tracker returns low-confidence outputs or struggles to keep the real-time performance.
This module is trained on video sequences ---it cannot exploit detection-only data as in~\cite{Zhu2018a}--- and its weights are adjusted alternatingly with the rest of the tracker, as the the MLP inputs must come from the RPN proposals.

\begin{figure}[!t]
\centering
\includegraphics[width=0.83\linewidth]{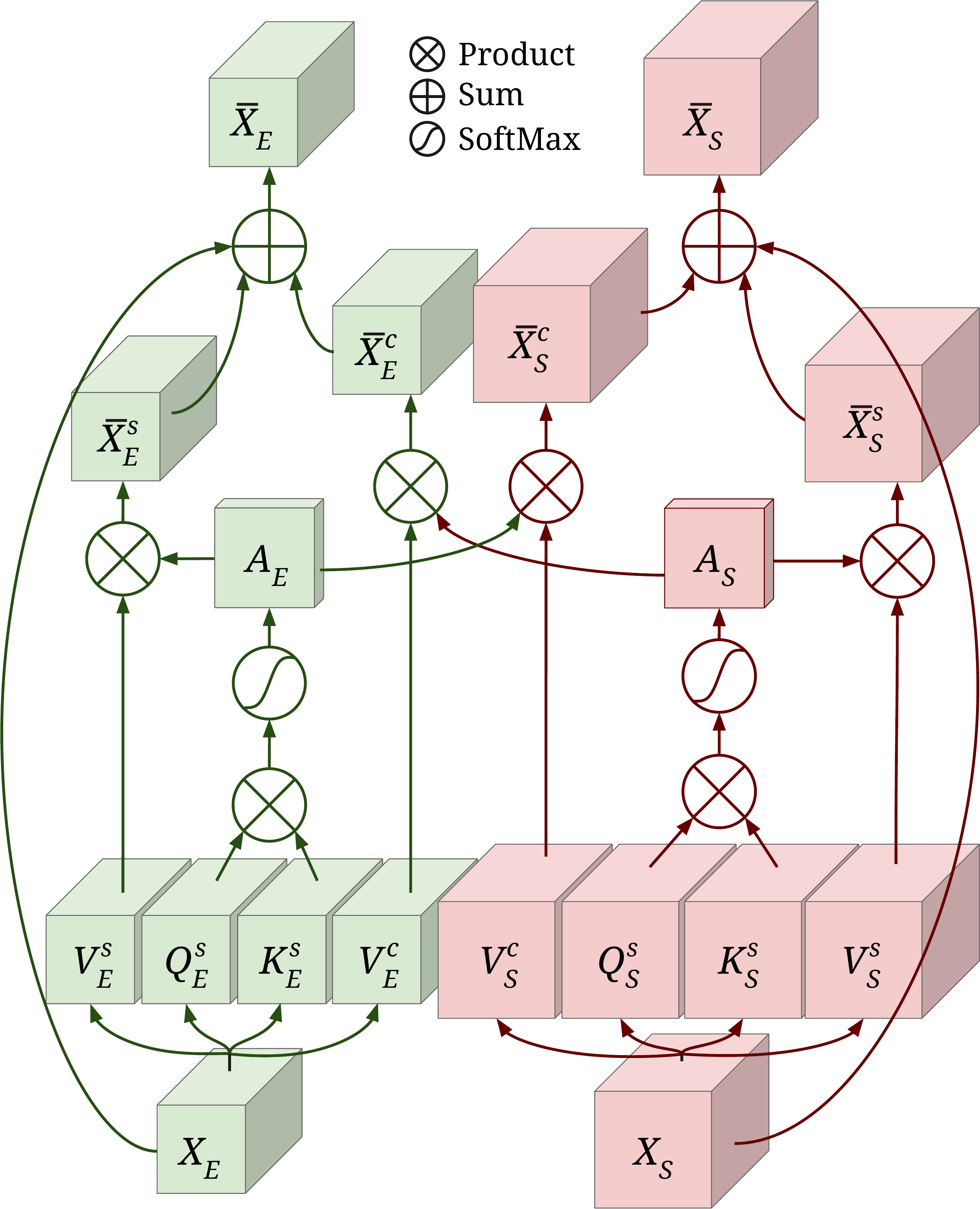}
\vspace*{-2mm}
\caption{\siammt's attention module has one branch for the {\textcolor{darkgreen}{\textbf{exemplar}}} and another one for the {\color{red} \textbf{search area}}.
Within each branch, channel-wise self-attention and cross-attention features are computed, which are then merged to obtain the exemplar and search area enhanced features.
The above operations are performed concurrently for each target.\vspace*{-1mm}}
\label{fig:attention}
\end{figure}

\subsubsection{Attention Module}
The similarity operator (Section~\ref{sec:similarity-operator}) performs a sliding-window comparison between the exemplar features of each object and the tensor containing its search area.
This very simple process is prone to problems, as the tracker has no way of knowing which features of the exemplar are truly relevant and tends to weigh everything equally.
Thus, it is common to experience a degradation in tracking quality when the background is complex or contains distractors.
To solve these problems, \siammt integrates into its architecture an attention module with the objective of mitigating drifting and enhancing the features it relies on.
This module is based on~\cite{Yu2020b}, but we have adapted it for multiple objects, suppressing the most expensive operations, and incorporating some optimizations.

Most of the attention mechanisms currently employed are based on the self-attention operation introduced in~\cite{Vaswani2017} for natural language processing.
In short, this operation takes a tensor $X$ as input and projects it into three different spaces, yielding three tensors denoted Query ($X_Q$), Key ($X_K$) and Value ($X_V$) ---as in information retrieval, we can think of Query as the search term, Key as the information that defines each of the candidates, and Value as the content of those candidates.
Next, it computes the cosine similarity between Query and Key, resulting in an interrelationship matrix ($A$) with high values for those features of $X$ that are related and relevant to the current problem, and low values otherwise.
This attention mask then is applied over Value, generating a filtered tensor ($\bar{X}$) in which unnecessary details are suppressed and important information has been highlighted.

To apply this concept to visual object tracking, it is necessary to adapt the aforementioned operation.
In~\cite{Yu2020b} it is proposed a deformable subnetwork comprising 3 complementary mechanisms: a channel-wise self-attention, a spatial-wise self-attention, and a channel-wise cross-attention.
According to our experiments, each of these components has a very different impact on system performance when dealing with several dozens of targets.
Most notably, spatial-wise self-attention requires several additional convolutions and reshapes that culminate in the resource-expensive multiplication of two $(W*H)\times(W*H)$ matrices ---for an input tensor $X$ of size $W \times H \times C$---, which diminishes the speed of the entire tracker by as much as $20\%$.
For this reason, \siammt bases its attention module solely on the channel-wise self-attention and cross-attention operations, and customizes their architecture for a more efficient performance (\figurename~\ref{fig:attention}).

\siammt applies its attention module in parallel for all \N pairs of tensors obtained through the RoI extractor.
However, for the sake of simplicity, we will focus the explanation on the computations involved for a single object.
Thus, the attention module receives as inputs the exemplar (${X}_{E}^{}$) and search area (${X}_{S}^{}$) features of an object.
For the self-attention computation ---whose tensors are denoted with the superscript~$s$---, 3 tensors are built in each branch ---${Q}_{E}^{s}$, ${K}_{E}^{s}$, and ${V}_{E}^{s}$ for the exemplar; and ${Q}_{S}^{s}$, ${K}_{S}^{s}$, and ${V}_{S}^{s}$ for the search area---, which have the same dimensions as ${X}_{E}^{}$ and ${X}_{S}^{}$, respectively.
Next, the cosine similarity between Query and Key is computed ---this is obtained via the dot product of the tensors scaled by their magnitude---, and then a SoftMax is applied, restricting the values to the interval $[0, 1]$, thus creating the attention masks ${A}_{E}^{}$ and ${A}_{S}^{}$.
Lastly, ${A}_{E}^{}$ is multiplied with ${V}_{E}^{s}$, and ${A}_{S}^{}$ with ${V}_{S}^{s}$, yielding the self-attention features for the exemplar (${\bar{X}}_{E}^{s}$) and the search area (${\bar{X}}_{S}^{s}$).

Regarding the cross-attention computation ---whose tensors are denoted with the superscript~$c$---, Value tensors for the exemplar (${V}_{E}^{c}$) and the search area (${V}_{S}^{c}$) branches are generated from the input features.
The latter are simply multiplied with the attention masks of the opposite branch, resulting in the cross-attention features of exemplar (${\bar{X}}_{E}^{c}$) and search area (${\bar{X}}_{S}^{c}$).
Finally, the attention features of each branch are combined with the original tensors through a weighted sum, producing the final enhanced features (${\bar{X}}_{E}^{}$) and ${\bar{X}}_{S}^{}$, whose sizes are the same as ${X}_{E}^{}$ and ${X}_{S}^{}$, respectively.

To improve the performance and speed up the learning, \siammt relies on standard 2D convolutions, as the accuracy gain that deformable convolutions offer does not compensate for the drop in throughput when dealing with several dozens of objects.
In addition to this, the aggregation of $X$ with ${\bar{X}}_{}^{s}$ and ${\bar{X}}_{}^{c}$ is performed in a single step, maximizing the use of computing resources.
Regarding the exemplar branch, in \siammt, exemplar features (${X}_{E}^{}$) are extracted once and reused throughout the whole tracking process.
Hence, it is possible to cache the tensors ${X}_{E}^{}$, ${\bar{X}}_{E}^{s}$, ${V}_{E}^{c}$, and ${A}_{E}^{}$ to speed up the computations of ${\bar{X}}_{E}^{}$ and ${\bar{X}}_{S}^{c}$ for the subsequent frames of the sequence.
The latter, although it implies a higher memory consumption per object, results in a significant increase in tracking speed.

\subsection{Comparison Head}
Once the features of the search areas are available, it is necessary to compare them with those of the exemplars in order to know the new coordinates of the objects.
Networks such as \mbox{SiamMT}~\cite{Vaquero2021} or \mbox{SiamFC}~\cite{Bertinetto2016} employ a simple cross-correlation for determining the location of the targets and rely on multi-scale testing for determining their size, which is fairly straightforward, but does not allow detecting changes in aspect ratio.
Approaches such as \mbox{SiamRPN}~\cite{Li2018} or \mbox{SiamFC\texttt{+}\texttt{+}}~\cite{Xu2020} employ more sophisticated region proposal networks, but perform slowly for multiple objects and tend to produce noisy outputs when distractors are present.
We propose a comparison head (CH) that efficiently compares multiple pairs of exemplars and search areas, generating quality predictions in multiple object scenarios.
The head comprises a Pairwise Depthwise Region Proposal Network (PD-RPN) and a multi-object penalization module.

\subsubsection{Similarity Operation\label{sec:similarity-operator}}
\siammt requires a fast and accurate similarity operator, capable of comparing dozens of tensors in real time.
Previous works~\cite{Bertinetto2016,Vaquero2021} computed the similarity between the search area and the exemplar straight up through a cross-correlation, obtaining a two-dimensional score map stating the probability of the object being in each region of the search area.
This can only identify translations, so multi-scale testings are carried out to detect changes in size.
The method is far from optimal, as it cannot detect aspect ratio variations and requires to compare $\Scales*\N$ feature maps, where $\Scales$ and $\N$ are the number of considered scales and objects, respectively.

\begin{figure*}[!t]
\centering
\includegraphics[width=0.94\linewidth]{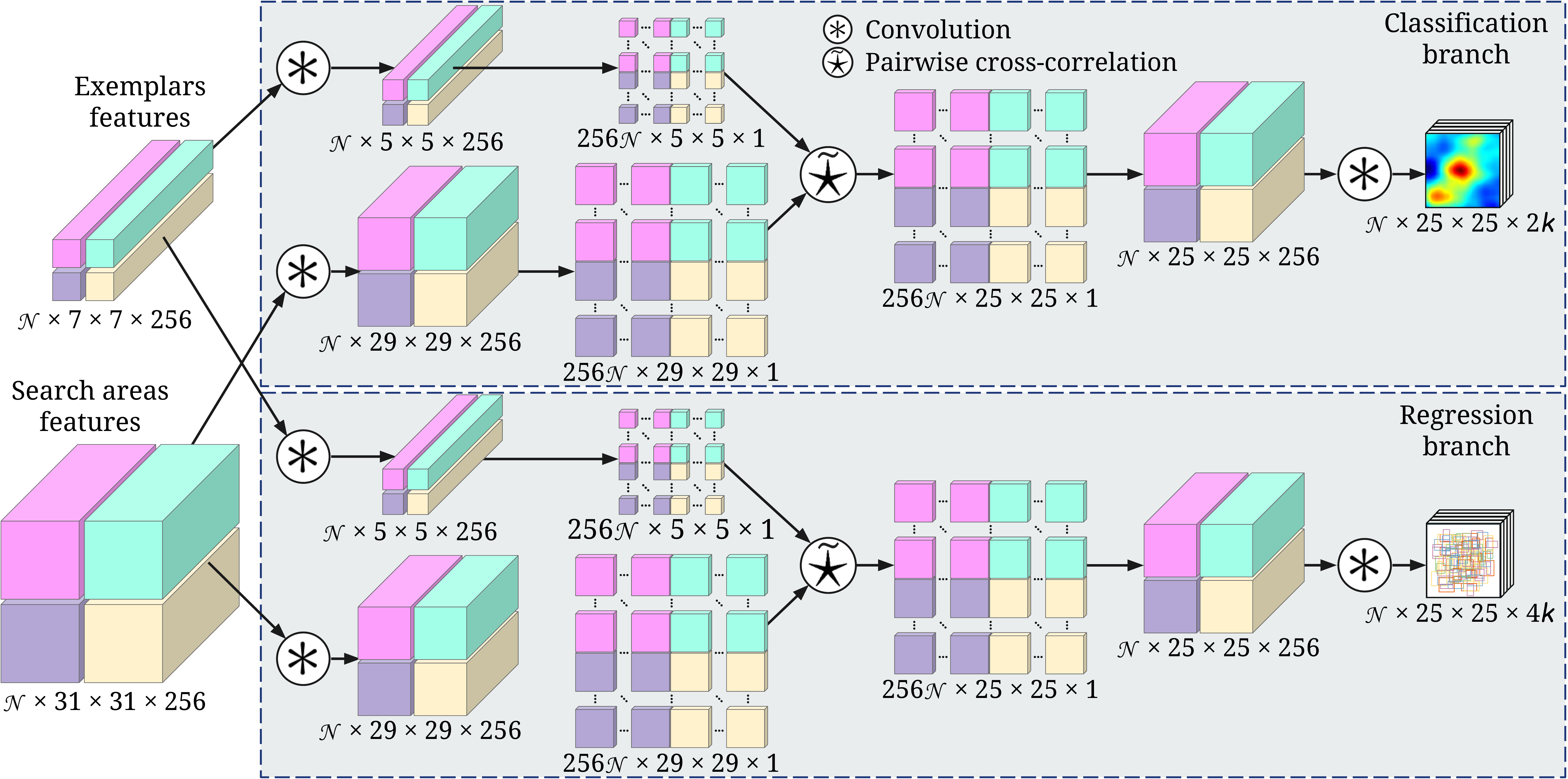}
\vspace*{-2mm}
\caption{Pairwise-Depthwise-RPN.
The input features pass through $3 \times 3$ convolutions that specialize them for each branch.
After this, they are reshaped to then perform a pairwise-depthwise-cross-correlation using \xcorrnew.
Finally, the features are reshaped a second time and $1 \times 1$ filters are applied to obtain $2K$ or $4K$ values per location ---depending on the branch---, yielding the classification and regression information of the different anchors.}
\label{fig:rpn}
\end{figure*}

To deal with changes in ratio and scale in a natural way, other methods such as~\cite{Li2018} employ a region proposal subnetwork.
They first transform the input features according to the number of considered anchors $K$ and then cross-correlate them, directly yielding the objects' classification and regression information.
The downside of this approach is that it requires $6K\N$ comparisons ---$2K\N$ for the objectness classification plus $4K\N$ for the bounding box regression---, which greatly increases the computational cost when there are many targets involved.
To address the aforementioned problems, \cite{Li2019} proposes a depthwise-RPN, which first cross-correlates the feature maps, to then apply the anchor-dependent transformations to their outputs.
This has the advantage of involving just $2\N$ comparisons, requiring an order of magnitude fewer parameters, and making the different channels more discriminative.

Owing to these benefits, \siammt implements its similarity operation based on the one introduced in~\cite{Li2019}.
However, as this operator is designed for single-object trackers, it is necessary to modify it as depicted in \figurename~\ref{fig:rpn} in order to enable it to process dozens of targets in real-time.
To achieve this, \siammt relies on the pairwise cross-correlation operator (\xcorrnew) described in~\cite{Vaquero2021} as the core of this Pairwise-Depthwise-RPN, which enables the computation of all objects and sub-windows in a single evaluation.
In order to perform these correlations in a depthwhise manner, it is however necessary to reshape the inputs and outputs of \xcorrnew.
Lastly, since the search areas are extracted through multi-level regions of interest, \siammt does not need an ensemble of heads at different stages of the backbone ---as \cite{Li2019} does---, which greatly improves the efficiency of the algorithm.

\subsubsection{Multi-Object Penalization Module}
\begin{figure}[!t] 
\centering
\includegraphics[width=0.90\linewidth]{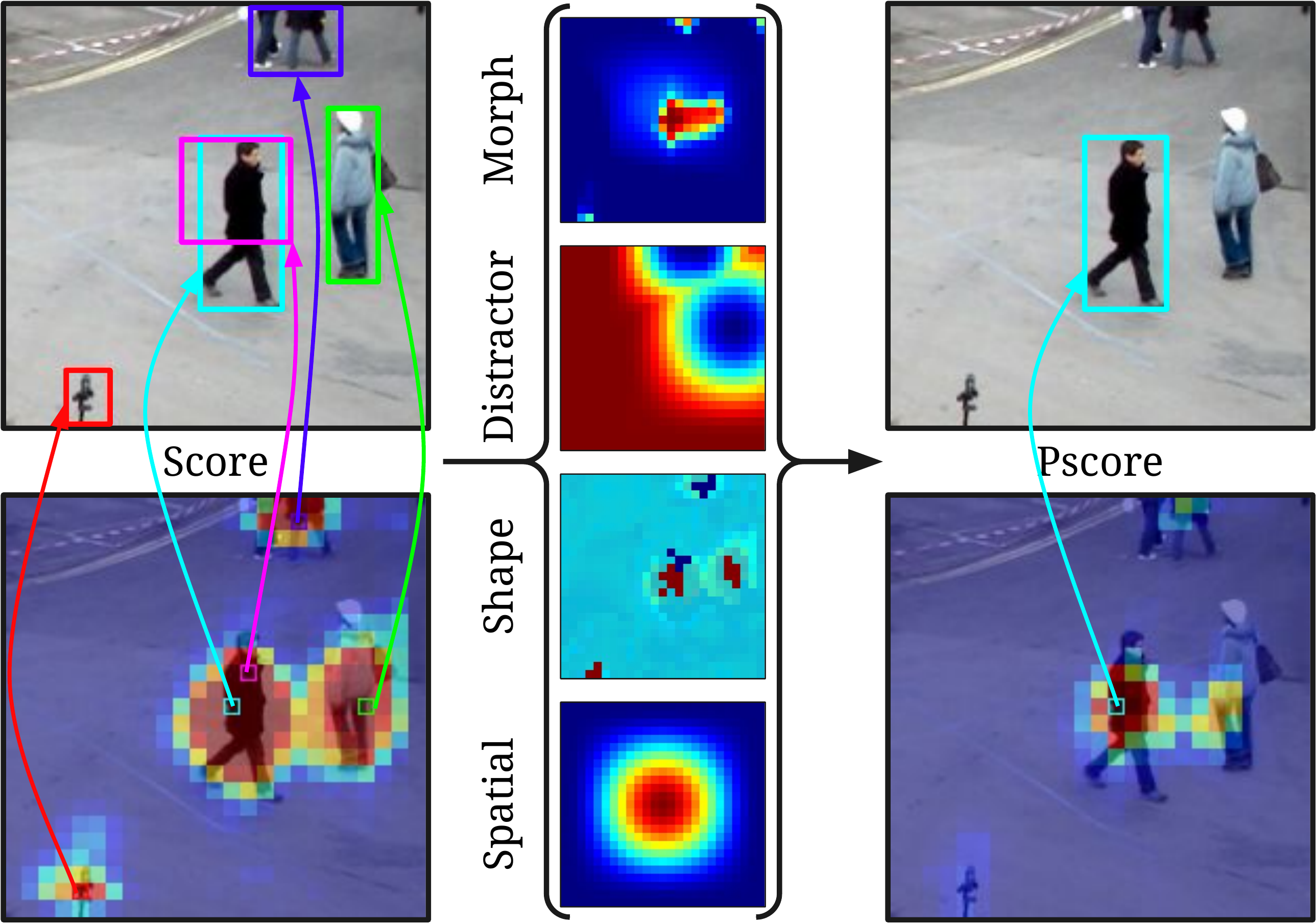} 
\caption{\siammt's multi-object penalization module.
The left side of the image depicts the output of the RPN classification branch with some of its associated bounding boxes, while the right contains the penalized score resulting from applying the four proposed penalties.}
\label{fig:penalizations}
\end{figure}

During inference, most single-object trackers~\cite{Bertinetto2016,Li2018,Bhat2019,Guo2020,Xu2020} refine their predictions using some form of heuristic knowledge.
The most commonly adopted methods are spatial and shape penalizations.
However, since \siammt is an MVOT and will consider several objects at once, it can exploit this information and develop a more sophisticated and powerful penalization module.
Thus, as shown in \figurename~\ref{fig:penalizations}, \siammt applies two new types of penalties ---distractor-aware and morphological---, resulting in a multi-object penalization module with four different types of refinements.
Additionally, the predictions produced by this component are fed to the inertia module, which uses them to roughly forecast the position of objects in future frames.

\textbf{Spatial penalization.}
Object velocities are usually moderate and situations in which a target undergoes a sudden large acceleration are extremely rare.
Thus, with a high probability, the object will remain in the central portions of the search area.
To model this knowledge, we follow the approach in~\cite{Bertinetto2016} and apply a two-dimensional Hanning window to the output of the classification branch in order to penalize large spatial displacements.
With this method, we are boosting proposals such as the blue, pink, and green found in \figurename~\ref{fig:penalizations}.

\textbf{Shape penalization.}
Whether it is due to changes in perspective or deformations, the objects in the scene change shape gradually in most cases.
Thus, to avoid accepting proposals with too different sizes or aspect ratios, we adopt a slightly modified version of the penalty described in~\cite{Li2018}:
\begin{equation}
\mathsf{\textit{penalty}} = e^{-\beta * \left(max\left(\frac{r}{r'}, \frac{r'}{r}\right) * max\left(\frac{s}{s'}, \frac{s'}{s}\right) - 1\right)}
\end{equation}
where $\beta$ is the hyper-parameter that adjusts the strength of the penalty, $r$ and $r'$ are the aspect ratios of the proposal and the object in the last frame, and $s$ and $s'$ are their overall scales, respectively.
This penalty, which is computed based on the regression branch and applied to the output of the classification branch, allows to suppress proposals which have a high objectness score but are linked to a poorly adjusted bounding box, such as the pink one in \figurename~\ref{fig:penalizations}.

\textbf{Distractor-aware penalization.}
As \siammt is an MVOT, it is aware of all the tracked objects that might potentially be similar to each other.
The proposed architecture takes advantage of this information and defines a novel distractor-aware penalization mask that minimizes identity-switches.
Thus, unlike other methods that handle each target independently, \siammt employs a simple-yet-effective approach to model interactions between objects.

The existence of distractors is modeled by a global penalization window~$G_{W \times H \times \N}$.
This is a tensor whose \N channels contain the contribution of each object to the distractor model, centered on their location and proportional to their size.
The window has a size of $(W, H) = \left \lceil \frac{M*F}{S} \right \rceil$ --- where $M$, $F$, and $S$ are the sizes of the RPN output classification map, the input frame, and the canonical search area size employed in Siamese trackers~\cite{Bertinetto2016}, respectively---, and the value of each element is computed as:
\begin{gather}\label{eq:global-window}
\begin{cases} 1 - \left(  \sin\left(\pi \frac{A_{p}^{x} - 0.5 A_{p}^{w} - i}{A_{p}^{w} - 1}\right)  \sin\left(\pi \frac{A_{p}^{y} - 0.5 A_{p}^{h} - j}{A_{p}^{y} - 1}\right)   \right)^{2}       , \ \ \ \ \ \ \ & \\
\multicolumn{2}{r}{\ \ \ \ \ \ \ \ \ \ \ \ \text{if $i \in A_{p}^{x} \pm \frac{A_{p}^{w}}{2}$ and $j \in A_{p}^{y} \pm \frac{A_{p}^{h}}{2}$}} \\
                     1, \ \ \   \text{otherwise} & \end{cases}
\raisetag{4.0mm}
\end{gather}
where $i \in [0, W)$, $j \in [0, H)$, $p \in [0, \N)$, and $A_{4 \times \N} = \left \{ (A_{p}^{x}, A_{p}^{y}, A_{p}^{w}, A_{p}^{h})\right \}_{p \in [0, \N)}$ is the objects' search area coordinates mapped over the penalization window.

From the model provided by $G_{W \times H \times \N}$, it is possible to estimate how the distractors will influence the model predictions.
Thus, to mitigate their influence, a window $R_p$ of size $M$ is computed for each object~$p$ as follows:
\begin{equation}\label{eq:global-window-per-target}
R_p = \min_{d \in \mathcal{D}} \; {\crop_{M}}{\left(G_{d}, A_p\right)}
\end{equation}
where $\mathcal{D} := \left \{\forall d \in [0, \N), d \ne p \right\}$ and $\crop_{M}$ is the preferred crop-and-resize operator with an output of size $M$ ---in our case, RoI Align~\cite{He2017}.
This yields a mask that can be applied to the probabilities of the classification branch, suppressing proposals such as the green and dark blue ones found in \figurename~\ref{fig:penalizations}.
This novel approach for modeling interactions between objects proves itself to be effective and has a low computational cost, which makes it ideal for environments with dozens of targets.

\textbf{Morphological penalization.}
When computing the similarity between two feature maps, there are some situations in which outliers appear.
These are characterized by being isolated false positives ---unlike correct matches, which are spread over a wide area--- and having a high score.
To suppress these points, we propose a novel penalization based on morphological operations. Specifically, we rely on erosion, whose main goal is to shrink the shapes contained in grayscale images.
Thus, we slide a $3\times3$ erosion kernel, which removes isolated peaks ---such as the red proposal in \figurename~\ref{fig:penalizations}--- and has the added benefit of reducing the area of the correct matches, as very wide boundaries can lead to problems in crowded scenarios.
We apply said window independently on each of the score map channels, as the activations for different anchor shapes are weakly related.

%% file: experimentation.tex

\section{Experiments\label{sec:experimentation}}
In this section, we assess the performance of \siammt under different scenarios.
The experiments were conducted on a computer with an Intel Core i7-9700K, 16~GB of DDR4 RAM and an NVIDIA TITAN Xp.
The chosen deep learning framework was TensorFlow.

\subsection{Implementation details}
\textbf{Feature extractor.}
We opted for ResNet~\cite{He2016} as it offers good results with low resource requirements, something to consider when including level $1$ of the FPN, which has a large memory footprint.
Regarding the specific network configuration, we chose ResNet-18, as it offers a good tradeoff between accuracy and speed.
It might seem that we can grow the backbone further without much impact on the throughput of the network since the following ResNet levels hardly increase the number of parameters.
However, according to our observations, FLOPS do not translate well to fps, thus using more complex backbones would make \siammt lose the ability to operate in real time.

\textbf{Inertia module.} The inertia module receives the information relative to the last $6$ known positions of the objects, since according to our experiments they are enough to generate reasonable predictions.
It comprises $2$ hidden fully-connected layers with hyperbolic tangent activations and ends in $4$ neurons with linear activations.
At the beginning of a sequence, the inputs of the MLP network are initialized to~$0$ ---as the module receives differences between coordinates, it simply assumes that the objects were stationary up to this point in time.
Regarding the training of the MLP, standard smooth $L_1$ loss is used for each component of $dB^{t}$.

\textbf{Exemplar and search area sizes.} The sizes of the exemplar and the search area influence many architecture decisions and, thus, deserve careful consideration.
Depending on their size, the output tensors will have a different granularity, which will directly impact the accuracy and speed of the network.
As proposed in \cite{Li2019}, an output of size $25\times25$ obtained from the comparison of exemplar and search area features of sizes $7\times7$ and $31\times31$, respectively, offers a good balance.

These tensors are created from the frame features for each object, using the RoI extractor.
Specifically, for a target with dimensions ($w$, $h$), its exemplar will cover an area of:
\begin{equation}\label{eq:area-calculation}
A^2 = \left(w + \zeta\left(w + h\right)\right) \times \left(h + \zeta\left(w + h\right)\right)
\end{equation}
where $\zeta=0.5$ is the context factor.
This area will be mapped to $15\times15$ bins.
However, since the cropping is performed on features, there is no need for extra context to accommodate for further convolution operations, so only the central $7\times7$ region is kept.
For the search area, a region of size $\left(\frac{31A}{15}\right)^2$ is cropped, which provides the same resolution as the exemplar.

\textbf{Training process.} \siammt is trained on a single GPU using the classification and regression losses defined in~\cite{Li2018} and employing an Adam optimizer~\cite{Kingma2015}, starting from a learning rate of $3 \times 10^{-5}$ that is exponentially decayed to $3 \times 10^{-7}$ with a batch size of $16$.
We build on a backbone pre-trained on ImageNet, which we freeze for the first $15$ epochs during the warmup stage.
Following this, we train the network end-to-end for $30$ epochs, applying a $0.1$ correction factor to the backbone gradients.
Lastly, we perform a fine-tune for $15$ epochs in which we favor samples drawn from video sequences and intra-class discrimination, and during which the inertia module is alternatingly trained with a dropout rate of $50\%$.
We maintain moving averages of the trained parameters with an exponential decay of $0.9998$, and add a weight decay of $5e-5$ to the loss function.

The system is trained on COCO\cite{Lin2014}, ILSVRC~\cite{Russakovsky2015}, YT-BB~\cite{Real2017}, and GOT-10k~\cite{Huang2019}, which comprise approximately $250$K~sequences and $1.5$M~images, for a total of roughly $11$M~bounding-boxes.
It receives pairs of images that will serve as exemplars and search areas, updating the weights of the network as it learns ---the feature extractor, RoI extractor, and attention module share their parameters for both exemplar and search area branches in a Siamese manner.
To speed up the training, the network is not fed with pairs of full frames, but with only those portions that contain objects.
These images undergo a data augmentation process that ensures their correct distribution into one of the levels of the FPN.
Another consideration to keep in mind is that the performance of the network will be severely harmed if the RoI Extractor is fed directly with the ground truth coordinates in those cases where the inertia module is not queried.
Therefore, it is very important to introduce noise in such circumstances, in addition to applying the spatial-aware sampling strategy described in~\cite{Li2019}.

\textbf{Inference process.} In order to be as efficient as possible, the exemplar features are extracted only once and reused throughout the rest of the inference process.
There is no need to update them, since they are already implicitly adapted for each frame thanks to the cross-attention mechanism.
The component of the network that is continuously updated is the inertia module, which is fed with the outputs of the predictor.
However, since it does not make use of visual information, it does not have a noticeable impact on the speed of the architecture.
Lastly, we apply non-maximum suppression and bounding box voting with an IoU of $80\%$ on the final output maps for a more precise localization.

\subsection{Ablation Study\label{sec:ablation}}
\siammt is an MVOT, which implies that it receives the initial position of each object and provides their locations in the following frames, with no further feedback from the detector.
These algorithms are mostly used in scenarios where time is a critical resource, so it must be taken into account when evaluating their performance.
As such, we evaluate the performance of \siammt through VOTChallenge's VOT-RT metrics~\cite{Kristan2019} adapting it for sequences containing multiple targets.
Thus, for a throughput threshold ---i.e., $20$~fps or $25$~fps--- we obtain the accuracy ---average overlap between predictions and ground truths--- and the robustness ---ratio of frames where the tracker did not lose the object, with an exponential sensitivity of $\gamma=30$--- of each algorithm.

In order to analyze the impact of each component of \siammt, we have conducted an ablation study on several scenarios in which motion estimation systems are commonly used.
Specifically, we have selected the multi-object public datasets:
\mbox{MOT-2017}~\cite{Milan2016a}, \mbox{MOT-2020}~\cite{Dendorfer2020}, \mbox{UAVDT}~\cite{Yu2020}, \mbox{VisDrone}~\cite{Zhu2020}, and \mbox{JTA}~\cite{Fabbri2018}.
The results are shown in Table~\ref{tab:benchmarks-ablation}.

\input{tables/benchmark-ablation.tex}

As the \textbf{baseline (B)} for our study, we build on an architecture very similar to~\cite{Vaquero2021}, but with a backbone based on ResNet-18~\cite{He2016} and without dynamically adjusting the input frame size.
The results are fairly good, but the performance is especially poor in videos with very small or very large objects, like the ones found in UAVDT, as the learned filters are not able to extract meaningful information.

The addition of the RoI extractor embedding a \textbf{feature pyramid network (F)} is a very significant change, increasing the accuracy and the robustness in all the datasets listed in Table~\ref{tab:benchmarks-ablation} by an average of $+4.1$ and $+5.8$ points, respectively.
The most substantial gain is found in UAVDT ---increase of~$+9.0$ points in accuracy and~$+16.7$ points in robustness---, as the network is now able to generate meaningful features at different resolutions and, therefore, handle small objects, which are very abundant in this dataset.

The similarity operator also has a major impact on network performance, as the change from a pairwise cross-correlation to our \textbf{pairwise-depthwise-RPN (R)} results in an average increase of~$+2.9$ points in accuracy.
This is mainly due to the fact that the system is now able to detect changes in the aspect ratio of objects, all while keeping the real-time performance.
The average increase in robustness is also significant ($+0.6$ points), since the output of the pairwise-depthwise-RPN objectness branch is very similar to that of pairwise cross-correlation, with both focusing on locating the center of the targets.

The integration of the \textbf{attention module (A)} results in an average increase of~$+0.2$ points in accuracy and~$+0.1$ points in robustness.
As many scenes contain multiple objects with a high degree of overlap ---which makes the center of the objects difficult to identify--- the attention mechanism focuses on producing richer features capable of better delimiting the boundaries of each object.
There are certain datasets in which one attention mechanism performs better than the combination of both.
However, the best overall results are obtained through the use of both approaches.

The addition of the \textbf{multi-object penalization module (P)} also provides a significant improvement in the performance of the network.
The suppression of distractors and outliers as well as the reduction of the activation area of detections provides clean results for the bounding box voting, which results in an average increase of~$+0.9$ points in accuracy and~$+0.3$ points in robustness.

\begin{figure}[!t] 
\centering
\includegraphics[width=0.90\linewidth]{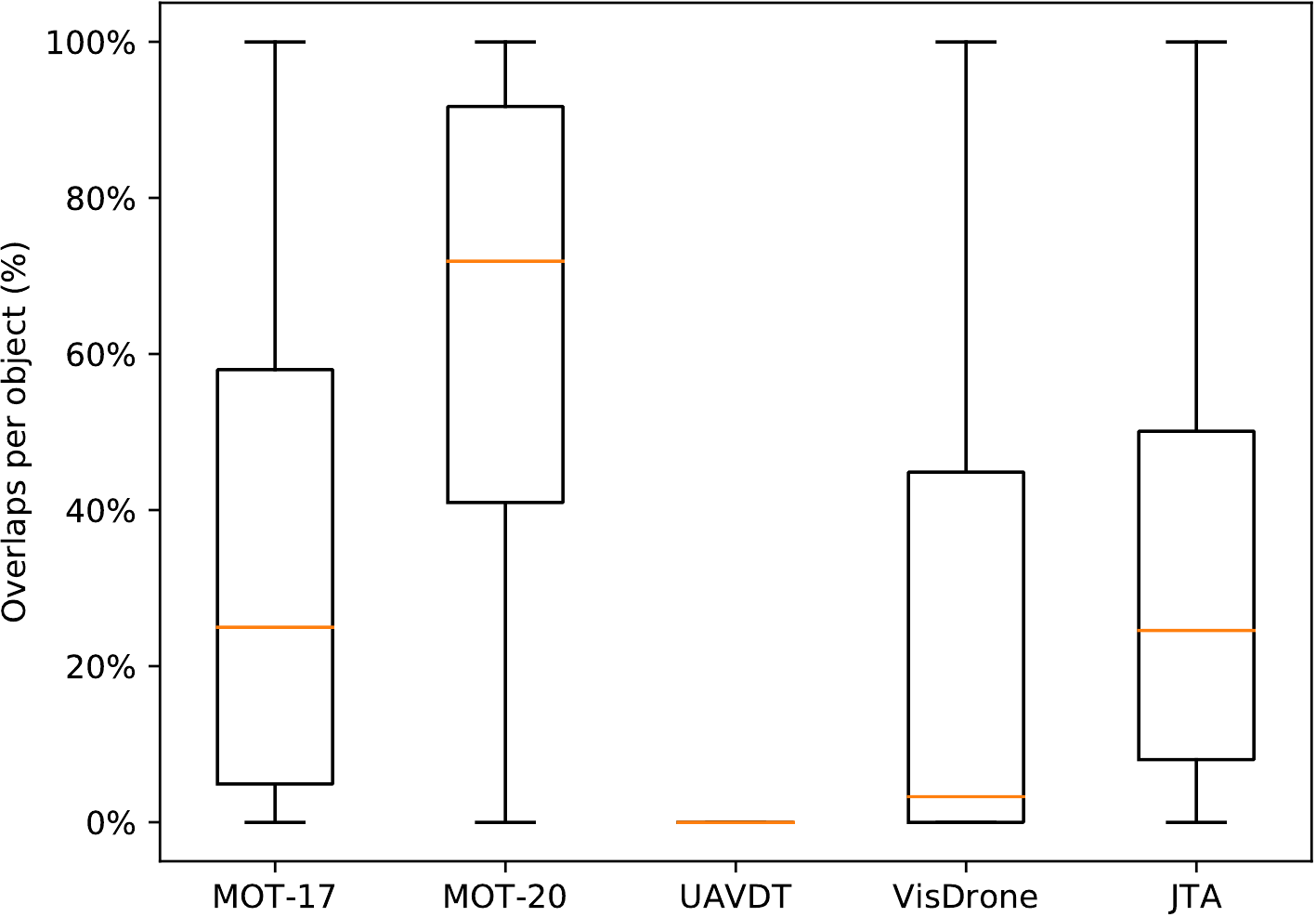}
\vspace*{-2mm}
\caption{Box plot of the amount of overlap per object for the analyzed databases. An object in a frame is regarded as overlapped if more than $50\%$ of its area is shared with other bounding boxes.\vspace*{-2mm}}
\label{fig:overlaps}
\end{figure}

Finally, the inclusion of the \textbf{inertia module (I)} also provides an improvement in the performance of the network, increasing the accuracy in~$+1.2$ points and the robustness in~$+0.5$ points for \mbox{MOT-2020} and \mbox{VisDrone}, respectively.
Specifically, as the inertia module is able to roughly predict the future coordinates of objects, it allows a better placement of the search areas, which prevents faster targets from leaving the network's field of view from one frame to the next.
On the other hand, in environments with a high density of objects where the visual information cannot be fully trusted due to continuous occlusions, the inertia module plays a major role in enhancing those RPN predictions that present low confidence.
Although the other datasets do not feature these characteristics, \mbox{VisDrone} has many fast targets and \mbox{MOT-2020} presents a large number of overlaps between objects (\figurename~\ref{fig:overlaps}), so the introduction of the inertia module in these cases offers a great advantage.

Overall, the proposed components improve the baseline by an average of~$+8.4$ points in accuracy and~$+6.8$ points in robustness.
The components that comprise the proposal engine add~$+4.6$ points in accuracy and~$+5.8$ points in robustness.
The comparison head is responsible for the increase of~$+3.8$ and~$+1.0$ points in accuracy and robustness, respectively.

These contributions not only improve the tracking quality, but they also preserve the efficiency of the algorithm.
The computational complexity of \siammt is of $\bigO(WH)$ for the feature extractor, $\bigO(WH+\N)$ for the RoI extractor, $\bigO(\N)$ for the inertia module, $\bigO(\N)$ for the attention module, $\bigO(K\N)$ for the pairwise-depthwise-RPN, and $\bigO({\N}^{2})$ for the multi-object penalization.
All this yields a computational complexity of $\bigO(WH+K\N+{\N}^{2})$ for the end-to-end network.
If we take into account that $W \simeq H \gg \N$, then the complexity is dominated by the size of the input frame.

Lastly, although the proposed architecture is already quite lightweight, it can be further simplified for deployment in industrial environments with constrained resources.
The core of the proposal engine is the region-of-interest extractor \textbf{(F)}, which leverages a feature pyramid network to yield features with richer semantic information.
On the other hand, the main component of the comparison head is the similarity operator \textbf{(R)}, which exploits a pairwise cross-correlation to fuse the exemplar and search area tensors while scaling seamlessly with the number of targets.
This configuration (Table~\ref{tab:benchmarks-ablation}, row ``\textbf{B+F+R}''), while simple, still outperforms the state of the art in all of the tested benchmarks but \mbox{MOT-2020}~\cite{Dendorfer2020} by an average of $+2.7$ points in accuracy and $+2.9$ points in robustness.
Nonetheless, the additional methods introduced in the paper are still beneficial for achieving superior performance, especially in crowded environments such as \mbox{MOT-2020}.

\input{tables/benchmark-tracking-quality-MOTRT.tex}

\subsection{Comparison with the state of the art\label{sec:experimentation_quality}}

Since \siammt is an MVOT, it constitutes a dedicated component that performs a specific task within the MOT framework ---motion estimation when there are no detections available.
Therefore, in this section, we compare it with current MVOT solutions on the previously discussed databases, employing VOTChallenge's VOT-RT metric @20fps and @25fps~\cite{Kristan2019}.
Specifically, this comparison covers \mbox{SiamFC}~\cite{Bertinetto2016} ---used in \mbox{UMA}~\cite{Yin2020}---, \mbox{SiamRPN}~\cite{Li2018} ---used in \mbox{DeepMOT}~\cite{Xu2020c} and \mbox{DAMOT}~\cite{Zhou2020a}---, and \mbox{SiamRPN\texttt{+}\texttt{+}}~\cite{Li2019} ---used in \mbox{SiamMOT}~\cite{Shuai2021}---, as well as \mbox{SiamMT}~\cite{Vaquero2021} and current state-of-the-art single-object trackers that can be tailored for MVOT ---\mbox{DaSiamRPN}~\cite{Zhu2018a}, \mbox{SiamCAR}~\cite{Guo2020}, \mbox{SiamFC\texttt{+}\texttt{+}}~\cite{Xu2020}, \mbox{SiamAttn}~\cite{Yu2020b}, \mbox{STMTrack}~\cite{Fu2021}, and \mbox{SiamBAN}~\cite{Chen2022}.
The results for each algorithm are shown in Table~\ref{tab:benchmarks-quality-MOTRT}.
All of these MVOTs have been run on multiple-object environments ---as the aforementioned papers do--- and maximizing GPU usage.
Column \textbf{\mypound{}ob} shows the maximum number of targets that can be instantiated before GPU memory oversubscription.

According to the experimental results, \siammt outperforms previous state-of-the-art MVOT approaches.
These improvements are in both accuracy and robustness, resulting in fewer identity switches and better delimiting the tracked objects.
Some approaches such as \mbox{SiamFC\texttt{+}\texttt{+}} are able to fit the bounding boxes to the objects very tightly, but have problems maintaining their identities for several frames ---i.e. high accuracy but low robustness.
Conversely, other approaches such as \mbox{DaSiamRPN} exhibit the opposite behavior.
These, while correctly identifying the center of the targets, report a bounding box with an incorrect size often vaguely related with the objects ---i.e. high robustness, but low accuracy.
The latter, while certainly not ideal, may be acceptable in those scenarios in which the boundaries of the objects are not a critical factor, but the main objective is to maintain their identities.

Owing to the global frame features extraction and the use of a specialized similarity operator (\xcorrnew), \mbox{SiamMT} fares better than its predecessors.
Consequently, our approach \siammt exploits these breakthroughs and builds on them to create a novel architecture.
Thus, the global features extraction allows to share the backbone computations ---the most expensive operation in the network---, and we incorporate the \xcorrnew operator as the core of our pairwise-depthwise-RPN.
This, together with the enhanced RoI extractor, the multi-object penalization module, and the inertia system, gives SiamMOTION advantages of up to~$+5.2$ points of accuracy and~$+7.6$ points of robustness compared to its predecessor.
When compared to the second best state-of-the-art architecture, these differences can become as large as~$+23.2$ and~$+25.1$ points, respectively.

One of the datasets in which \siammt best performs is \mbox{UAVDT} ---improvement of~$+3.2$ points in accuracy and~$+1.0$ points in robustness @20~fps.
This is because it contains a large number of small moving vehicles, for which the inertia module and the inclusion of the FPN in the RoI extractor are ideal.
The inertia module allows for better placement of the search area when there are fast movements ---this is critical for small objects, as the field-of-view of the network is smaller for them (\equationautorefname~\ref{eq:area-calculation})---, while the FPN provides meaningful features for small objects, which would normally be washed-out at deeper levels of the backbone.
Other datasets in which \siammt excels are \mbox{VisDrone} ---improvement of~$+5.2$ points in accuracy and~$+7.6$ points in robustness @20~fps--- and \mbox{JTA} ---improvement of~$+4.1$ points in accuracy and~$+4.8$ points in robustness @20~fps.
The large number of objects they contain and the fact that the sequences are captured with a moving camera mean that the bounding boxes of the targets will continuously change in aspect ratio.
Thanks to \siammt's built-in \mbox{PD-RPN}, we are able to detect these changes in an efficient and effective manner.

On the most challenging dataset, \mbox{MOT-2020}, \siammt achieves the highest difference in accuracy and robustness with the state of the art but SiamMT, surpassing them by over~$+21.5$ points in accuracy and~$+9.2$ points in robustness @20~fps.
This dataset is quite complex due to the large amount of overlap between objects and erratic movements it contains, making it completely different from the rest of the benchmarks ---the initializations themselves often contain parts of other tracked objects.
In fact, \figurename~\ref{fig:overlaps} shows how half of the objects in \mbox{MOT-2020} present overlaps in $70\%$ or more of their detections.
This statistic is very different from the rest of the databases, in which most of their objects are overlapped in less than $25\%$ of their detections.
An extreme example of this is \mbox{UAVDT}, where $90\%$ of its objects have an overlap below $3\%$, since it consists of zenithal recordings of vehicles.

%% file: tables/benchmark-ablation.tex

\begin{table*}[!t]
\centering
\caption{Ablation study with VOT-RT metrics @20 fps. A version of \mbox{SiamMT} with a ResNet-18 as backbone is the baseline (B). F: Feature-Pyramid-based RoI extractor. R: PD-RPN similarity operation. A: Attention module ---superscripts s and c indicate self-attention or cross-attention only, respectively. P: Multi-object penalization. I: Inertia module. The number of parameters is shown together with the name of each configuration.}
\label{tab:benchmarks-ablation}
\begin{tabular}{l@{\hskip 4pt}c@{\hskip 2pt}c@{\hskip 6pt}c@{\hskip 2pt}c@{\hskip 6pt}c@{\hskip 2pt}c@{\hskip 6pt}c@{\hskip 2pt}c@{\hskip 6pt}c@{\hskip 2pt}c}
\hline
\multicolumn{1}{c@{\hskip 4pt}}{\textbf{}} & \multicolumn{2}{c@{\hskip 4pt}}{\textbf{\hskip -11pt MOT-17}} & \multicolumn{2}{c@{\hskip 4pt}}{\textbf{\hskip -11pt MOT-20}} & \multicolumn{2}{c@{\hskip 4pt}}{\textbf{\hskip -11pt UAVDT}} & \multicolumn{2}{c@{\hskip 4pt}}{\textbf{\hskip -11pt VisDrone}} & \multicolumn{2}{c}{\textbf{\hskip -11pt JTA}} \vspace*{-1pt} \\
\textbf{} & Acc. & Rob. & Acc. & Rob. & Acc. & Rob. & Acc. & Rob. & Acc. & Rob. \\ \hline
\textbf{Baseline}~(13.9M) & 50.1 & 74.6 & 47.9 & 82.1 & 42.0 & 75.8 & 42.8 & 50.2 & 45.3 & 62.1 \\
\textbf{B+F}~(15.2M) & 55.4 & 76.8 & 48.1 & 82.3 & 51.0 & 92.5 & 46.3 & 56.0 & 48.0 & 66.1 \\
\textbf{B+F+R}~(16.4M)       & 57.3 & 76.6 & 51.2 & 83.0 & 56.5 & 93.2 & 48.7 & 57.0 & 49.8 & 67.1 \\
\textbf{B+F+R+A\textsuperscript{s}}~(17.0M)     & 57.4 & 76.5 & 50.8 & 83.3 & 56.6 & 93.1 & 48.9 & 57.2 & 49.9 & 67.1 \\
\textbf{B+F+R+A\textsuperscript{c}}~(17.0M)     & 57.2 & 76.7 & 51.8 & 83.2 & 57.2 & 93.2 & 48.7 & 57.1 & 49.8 & 67.1 \\
\textbf{B+F+R+A}~(17.0M)     & 57.3 & 76.6 & 51.4 & 83.2 & 57.0 & 93.2 & 48.9 & 57.1 & 50.0 & 67.2 \\
\textbf{B+F+R+A+P}~(17.0M)   & 58.0 & 77.5 & 51.5 & 82.7 & 57.8 & 93.2 & 50.8 & 59.3 & 51.0 & 66.2 \\
\textbf{B+F+R+A+P+I}~(17.0M) & 57.9 & 77.2 & 52.7 & 82.2 & 57.8 & 93.3 & 50.9 & 59.8 & 51.0 & 66.3 \\ \hline
\end{tabular}
\end{table*}

%% file: tables/benchmark-tracking-quality-MOTRT.tex

\begin{table*}[!t]
\centering
\caption{VOT-RT metrics results. {\color{red} \textbf{Red}}, {\color{blue} \textit{blue}} and {\color{green} green}, represent 1st, 2nd and 3rd respectively. Column \textbf{\mypound{}ob} shows the maximum number of targets that can be instantiated before the tracker resorts to GPU oversubscription.}
\label{tab:benchmarks-quality-MOTRT}
\resizebox{\linewidth}{!}{%
\begin{tabular}{@{\hskip 0pt}l@{\hskip 1pt}c@{\hskip 2pt}c@{\hskip 3pt}c@{\hskip 2pt}c@{\hskip 4pt}c@{\hskip 2pt}c@{\hskip 3pt}c@{\hskip 2pt}c@{\hskip 4pt}c@{\hskip 2pt}c@{\hskip 3pt}c@{\hskip 2pt}c@{\hskip 4pt}c@{\hskip 2pt}c@{\hskip 3pt}c@{\hskip 2pt}c@{\hskip 4pt}c@{\hskip 2pt}c@{\hskip 3pt}c@{\hskip 2pt}c@{\hskip 2pt}r@{\hskip 0pt}}
\hline
 & \multicolumn{4}{c}{\textbf{MOT-17}} & \multicolumn{4}{c}{\textbf{MOT-20}} & \multicolumn{4}{c}{\textbf{UAVDT}} & \multicolumn{4}{c}{\textbf{VisDrone}} & \multicolumn{4}{c}{\textbf{JTA}} & \vspace*{-2pt} \vspace*{-1pt} \\ 
\textbf{} & \multicolumn{2}{c}{@20 fps} & \multicolumn{2}{c}{@25 fps} & \multicolumn{2}{c}{@20 fps} & \multicolumn{2}{c}{@25 fps} & \multicolumn{2}{c}{@20 fps} & \multicolumn{2}{c}{@25 fps} & \multicolumn{2}{c}{@20 fps} & \multicolumn{2}{c}{@25 fps} & \multicolumn{2}{c}{@20 fps} & \multicolumn{2}{c}{@25 fps} & \vspace*{-2pt} \vspace*{-1pt} \\ 
\textbf{} & Acc. & Rob. & Acc. & Rob. & Acc. & Rob. & Acc. & Rob. & Acc. & Rob. & Acc. & Rob. & Acc. & Rob. & Acc. & Rob. & Acc. & Rob. & Acc. & Rob. & \textbf{\mypound{}ob} \\ \hline
\textbf{SiamMT}~\cite{Vaquero2021} & {\color{blue} \textit{54.5}} & {\color{blue} \textit{76.2}} & {\color{blue} \textit{54.5}} & {\color{blue} \textit{73.8}} & {\color{blue} \textit{52.6}} & {\color{blue} \textit{81.7}} & {\color{blue} \textit{52.5}} & {\color{blue} \textit{82.2}} & {\color{blue} \textit{54.6}} & {\color{blue} \textit{92.3}} & {\color{blue} \textit{54.5}} & {\color{blue} \textit{92.3}} & {\color{blue} \textit{45.7}} & {\color{blue} \textit{52.2}} & {\color{blue} \textit{45.4}} & {\color{blue} \textit{51.5}} & {\color{blue} \textit{46.9}} & {\color{blue} \textit{61.5}} & {\color{blue} \textit{46.9}} & {\color{blue} \textit{61.5}} & {\color{red} \textbf{$\bm{{>} {99}}$}} \\
\textbf{SiamBAN}~\cite{Chen2022} & 45.9 & 59.9 & 44.2 & 58.4 & 27.5 & 67.4 & 26.7 & 66.9 & 38.2 & 67.1 & 35.1 & 62.3 & 26.8 & 29.7 & 27.3 & 29.2 & 36.3 & 48.5 & 34.8 & 46.8 & $44$ \\
\textbf{STMTrack}~\cite{Fu2021} & 48.7 & 61.8 & 46.6 & 59.9 & 27.3 & 67.3 & 26.9 & 66.9 & 35.6 & 62.8 & 33.4 & 59.2 & 28.1 & 29.5 & 27.0 & 29.2 & 38.2 & 49.5 & 36.6 & 47.6 & $23$ \\
\textbf{SiamAttn}~\cite{Yu2020b} & 45.0 & 57.3 & 43.9 & 56.1 & 26.7 & 66.7 & 26.3 & 66.5 & 35.0 & 60.9 & 32.7 & 57.4 & 26.8 & 29.0 & 26.2 & 28.7 & 35.0 & 45.7 & 33.9 & 44.7 & $21$ \\
\textbf{SiamFC\texttt{+}\texttt{+}}~\cite{Xu2020} & {\color{green} {53.9}} & 70.2 & {\color{green} 52.3} & 67.6 & 30.4 & 70.9 & 28.8 & 69.7 & 52.4 & 86.1 & 48.5 & 81.2 & 34.2 & 32.8 & 32.0 & 31.5 & {\color{green} 44.4} & 59.8 & {\color{green} 42.1} & 56.2 & $30$ \\
\textbf{SiamCAR}~\cite{Guo2020} & 47.5 & 60.0 & 46.0 & 58.2 & 27.8 & 67.5 & 27.2 & 67.0 & 40.3 & 70.0 & 37.1 & 64.9 & 28.9 & 29.8 & 27.8 & 29.3 & 38.3 & 48.7 & 36.6 & 46.9 & {\color{green} $45$} \\
\textbf{SiamRPN\texttt{+}\texttt{+}}~\cite{Li2019} & 48.5 & 60.1 & 47.0 & 58.9 & 27.8 & 67.4 & 27.1 & 66.9 & 38.2 & 67.1 & 35.2 & 62.5 & 28.2 & 29.6 & 27.2 & 29.1 & 37.8 & 48.7 & 36.1 & 46.9 & $22$ \\
\textbf{DaSiamRPN}~\cite{Zhu2018a} & 48.2 & {\color{green} {72.5}} & 46.9 & {\color{green} {69.3}} & {\color{green} {31.2}} & {\color{green} {73.0}} & {\color{green} {29.5}} & {\color{green} {71.4}} & 52.0 & 89.5 & 48.6 & 85.4 & 35.3 & {\color{green} {34.7}} & 33.7 & {\color{green} {33.1}} & 42.0 & {\color{green} {61.6}} & 40.54 & {\color{green} {58.8}} & $33$ \\
\textbf{SiamRPN}~\cite{Li2018} & 50.9 & 70.7 & 49.5 & 67.9 & 30.3 & 71.0 & 29.1 & 69.7 & {\color{green} 53.5} & {\color{green} 89.9} & {\color{green} 49.8} & {\color{green} 85.7} & {\color{green} 36.2} & 34.4 & {\color{green} 33.9} & 32.8 & 43.7 & 60.3 & 41.5 & 56.8 & $33$
\\
\textbf{SiamFC}~\cite{Bertinetto2016} & 45.8 & 58.4 & 45.2 & 57.2 & 26.9 & 67.3 & 26.6 & 66.9 & 41.4 & 73.3 & 37.5 & 67.3 & 27.9 & 29.5 & 27.0 & 29.0 & 36.3 & 46.3 & 35.2 & 45.1 & $39$ \\ \hline

\textbf{SiamMOTION} & {\color{red} \textbf{57.9}} & {\color{red} \textbf{77.2}} & {\color{red} \textbf{57.8}} & {\color{red} \textbf{76.7}} & {\color{red} \textbf{52.7}} & {\color{red} \textbf{82.2}} & {\color{red} \textbf{52.7}} & {\color{red} \textbf{82.2}} & {\color{red} \textbf{57.8}} & {\color{red} \textbf{93.3}} & {\color{red} \textbf{57.7}} & {\color{red} \textbf{93.4}} & {\color{red} \textbf{50.9}} & {\color{red} \textbf{59.8}} & {\color{red} \textbf{49.2}} & {\color{red} \textbf{57.6}} & {\color{red} \textbf{51.0}} & {\color{red} \textbf{66.3}} & {\color{red} \textbf{51.0}} & {\color{red} \textbf{66.3}} & {\color{red} \textbf{$\bm{{>} {99}}$}} \\ \hline
\end{tabular}}
\end{table*}

%% file: conclusions.tex

\section{Conclusions and future work\label{sec:conclusions}}
We have presented \siammt, an MVOT (multiple visual object tracker) capable of tracking several dozens of objects in real-time with high accuracy and robustness, regardless of their category and size.
This is made possible thanks to a proposal engine that generates quality features ---well-framed, with the correct resolution, and highlighting the most relevant channels for each object--- and a comparison head that efficiently outputs quality predictions ---detecting changes in aspect ratio and suppressing the effect of distractors, all without resorting to multi-scale testing.

\siammt has been evaluated on various video databases, achieving a real-time performance that surpasses the current state of the art ---increase of~$+2.9$ points in both average accuracy and average robustness when compared to the best performing counterpart at $25$~fps, and differences of more than~$+11.6$ points when compared to the rest.
Furthermore, an exhaustive ablation study was carried out on all the tested databases, analyzing the contribution of each new component.
This showed that the parts comprising the proposal engine ---inertia module, RoI extractor, and attention mechanism--- are responsible for the $68\%$ of the improvement brought by the architecture ---$+4.6$ points in accuracy and~$+5.8$ points in robustness over the baseline---, while the components that make up the comparison head ---Pairwise Depthwise RPN and multi-object penalization module--- contribute the remaining $32\%$ ---$+3.8$ and~$+1.0$ points in accuracy and robustness, respectively.

As \siammt is an MVOT and not a fully-fledged MOT system, it solves a specific task within the MOT framework ---motion estimation when there are no detections available.
It is therefore not responsible for carrying out other tasks such as track initialization and termination, or drift detection.
This is why as future work it would be interesting to develop a complete MOT system integrating \siammt as well as an object detector, an affinity estimator, and an association mechanism; making it fully integrated and end-to-end trainable.
This would most likely allow for better learning for all components, as well as ease its deployment in resource-constrained environments.
It would also be interesting to transform \siammt to produce more thorough outputs, enabling rotated bounding boxes or per-pixel identification ---segmentation.
This would allow for more refined predictions, making tracking more valuable in those situations where objects have unusual shapes or appear in large crowds.
However, this is not straightforward, as such methods usually have a severe impact on the speed of the system, preventing it from running in real-time.

%% file: thanks.tex

\section*{Acknowledgment}
This research was partially funded by the Spanish Ministerio de Ciencia e Innovaci\'on {[grant numbers \mbox{PID2020-112623GB-I00}, \mbox{RTI2018-097088-B-C32}]}, and the Galician Conseller\'ia de Cultura, Educaci\'on e Universidade \hspace{0pt}[grant numbers \mbox{ED431C 2018/29}, \mbox{ED431C 2021/048}, \mbox{ED431G 2019/04}].
These grants are co-funded by the European Regional Development Fund (ERDF).
Lorenzo Vaquero is supported by the Spanish Ministerio de Universidades under the FPU national plan (FPU18/03174).
We also gratefully acknowledge the support of NVIDIA Corporation for hardware donations used for this research.

%% file: biography.tex

\par\noindent 
\parbox[t]{\linewidth}{
\noindent\parpic{\includegraphics[width=1in,height=1.25in,clip,keepaspectratio]{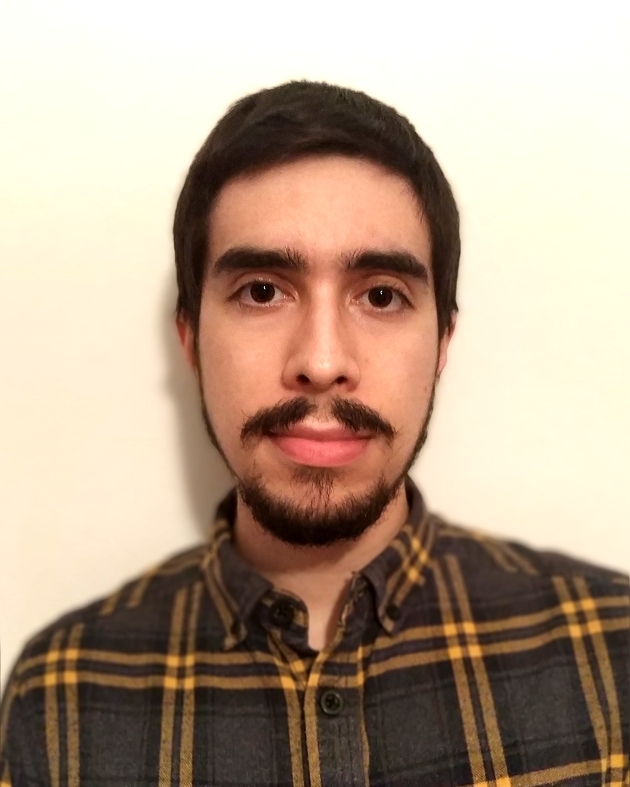}}
\noindent {\bf Lorenzo Vaquero}\
is a Ph.D. student at the CiTIUS of the University of Santiago de Compostela, Spain.
He received the B.S. degree in Computer Science in 2018 and the M.S. degree in Big Data in 2019.
His research interests are visual object tracking and deep learning for autonomous vehicles.}
\vspace{0.25\baselineskip}

\par\noindent 
\parbox[t]{\linewidth}{
\noindent\parpic{\includegraphics[width=1in,height=1.25in,clip,keepaspectratio]{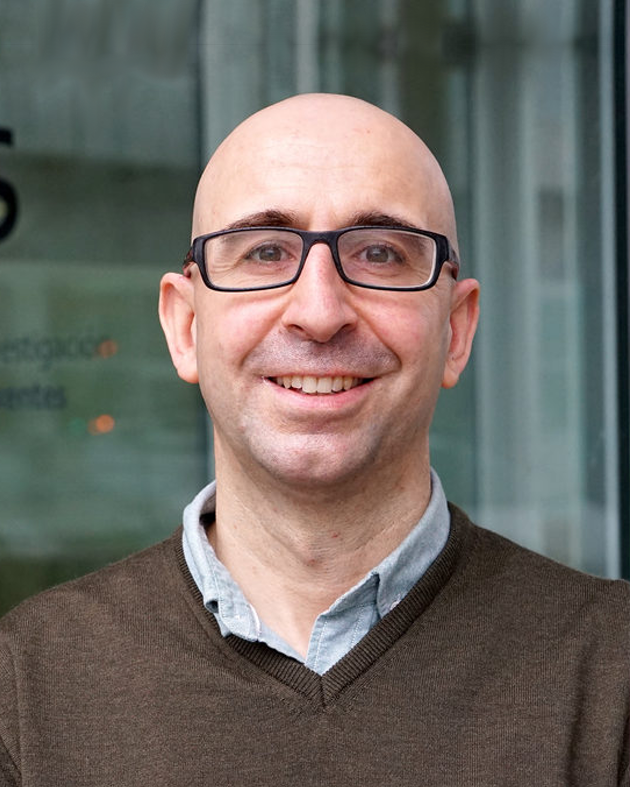}}
\noindent {\bf V\'ictor M. Brea}\
is an Associate Professor at CiTIUS, University of Santiago de Compostela, Spain.
His main research interest lies in Computer Vision, both on deep learning algorithms, and on the design of efficient architectures and CMOS solutions.
He has authored more than 100 scientific papers in these fields of research.}
\vspace{0.25\baselineskip}

\par\noindent 
\parbox[t]{\linewidth}{
\noindent\parpic{\includegraphics[width=1in,height=1.25in,clip,keepaspectratio]{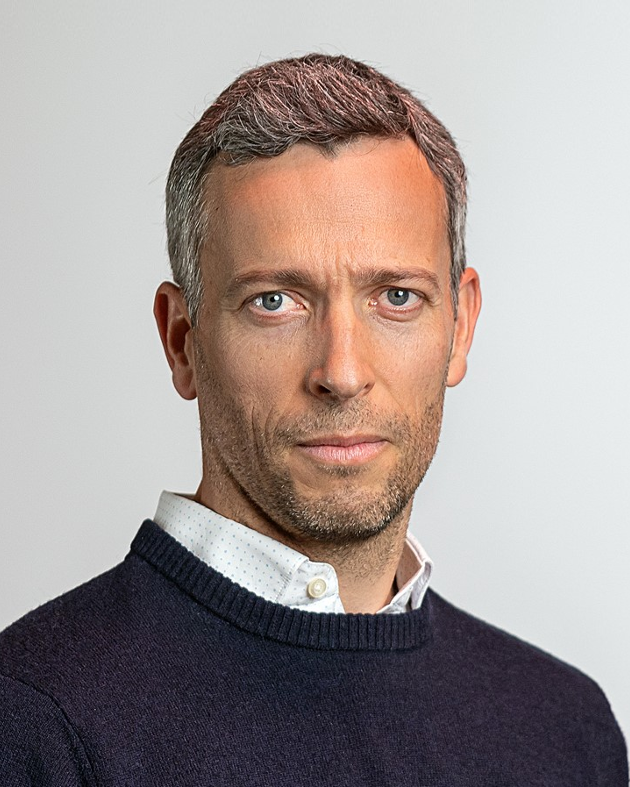}}
\noindent {\bf Manuel Mucientes}\
is an Associate Professor at the CiTIUS of the University of Santiago de Compostela, Spain.
His main research interest is artificial intelligence applied to the following areas: computer vision for object detection and tracking; machine learning; process mining.
He has authored more than 100 scientific papers in these fields of research.}

%% file: main.bbl
\begin{thebibliography}{10}
\expandafter\ifx\csname url\endcsname\relax
  \def\url#1{\texttt{#1}}\fi
\expandafter\ifx\csname urlprefix\endcsname\relax\def\urlprefix{URL }\fi
\expandafter\ifx\csname href\endcsname\relax
  \def\href#1#2{#2} \def\path#1{#1}\fi

\bibitem{Fang2021}
B.~Fang, G.~Mei, X.~Yuan, et~al., Visual {SLAM} for robot navigation in
  healthcare facility, Pattern Recognit. 113 (2021) 107822.

\bibitem{Dendorfer2020}
P.~Dendorfer, H.~Rezatofighi, A.~Milan, et~al., {MOT20:} {A} benchmark for
  multi object tracking in crowded scenes, CoRR abs/2003.09003.

\bibitem{Tan2020}
M.~Tan, R.~Pang, Q.~V. Le, Efficientdet: Scalable and efficient object
  detection, in: {IEEE} Conf. Comput. Vis. Pattern Recognit. ({CVPR}), 2020,
  pp. 10778--10787.

\bibitem{Vaquero2021}
L.~Vaquero, M.~Mucientes, V.~M. Brea, Tracking more than 100 arbitrary objects
  at 25 fps through deep learning, Pattern Recognit. 121 (2022) 108205.

\bibitem{Ciaparrone2020}
G.~Ciaparrone, F.~L. S{\'{a}}nchez, S.~Tabik, et~al., Deep learning in video
  multi-object tracking: {A} survey, Neurocomputing 381 (2020) 61--88.

\bibitem{Kristan2019}
M.~Kristan, J.~Matas, A.~Leonardis, et~al., The seventh visual object tracking
  {VOT}2019 challenge results, in: {IEEE} Int. Conf. Comput. Vis. ({ICCV})
  Workshops, 2019, pp. 2206--2241.

\bibitem{Fernandez-Sanjurjo2021}
M.~Fern\'{a}ndez-Sanjurjo, M.~Mucientes, V.~Brea, Real-time multiple object
  visual tracking for embedded {GPU} systems, {IEEE} Internet Things J. 8
  (2021) 9177--9188.

\bibitem{Bewley2016}
A.~Bewley, Z.~Ge, L.~Ott, et~al., Simple online and realtime tracking, in:
  {IEEE} Int. Conf. Image Process. ({ICIP}), 2016, pp. 3464--3468.

\bibitem{Zhou2020a}
Z.~Zhou, W.~Luo, Q.~Wang, et~al., Distractor-aware discrimination learning for
  online multiple object tracking, Pattern Recognit. 107 (2020) 107512.

\bibitem{Yin2020}
J.~Yin, W.~Wang, Q.~Meng, et~al., A unified object motion and affinity model
  for online multi-object tracking, in: {IEEE} Conf. Comput. Vis. Pattern
  Recognit. ({CVPR}), 2020, pp. 6767--6776.

\bibitem{Bolme2010}
D.~S. Bolme, J.~R. Beveridge, B.~A. Draper, Y.~M. Lui, Visual object tracking
  using adaptive correlation filters, in: {IEEE} Conf. Comput. Vis. Pattern
  Recognit. ({CVPR}), 2010, pp. 2544--2550.

\bibitem{Yuan2020}
D.~Yuan, X.~Li, Z.~He, et~al., Visual object tracking with adaptive structural
  convolutional network, Knowl. Based Syst. 194 (2020) 105554.

\bibitem{Xu2020a}
T.~Xu, Z.~Feng, X.~Wu, J.~Kittler, An accelerated correlation filter tracker,
  Pattern Recognit. 102 (2020) 107172.

\bibitem{Bertinetto2016}
L.~Bertinetto, J.~Valmadre, J.~F. Henriques, et~al., Fully-convolutional
  siamese networks for object tracking, in: European Conf. Comput. Vis.
  ({ECCV}) Workshops, 2016, pp. 850--865.

\bibitem{Li2018}
B.~Li, J.~Yan, W.~Wu, et~al., High performance visual tracking with siamese
  region proposal network, in: {IEEE} Conf. Comput. Vis. Pattern Recognit.
  ({CVPR}), 2018, pp. 8971--8980.

\bibitem{Guo2020}
D.~Guo, J.~Wang, Y.~Cui, et~al., Siamcar: Siamese fully convolutional
  classification and regression for visual tracking, in: {IEEE} Conf. Comput.
  Vis. Pattern Recognit. ({CVPR}), 2020, pp. 6268--6276.

\bibitem{Li2019}
B.~Li, W.~Wu, Q.~Wang, et~al., Siamrpn++: Evolution of siamese visual tracking
  with very deep networks, in: {IEEE} Conf. Comput. Vis. Pattern Recognit.
  ({CVPR}), 2019, pp. 4282--4291.

\bibitem{Yu2020b}
Y.~Yu, Y.~Xiong, W.~Huang, M.~R. Scott, Deformable siamese attention networks
  for visual object tracking, in: {IEEE} Conf. Comput. Vis. Pattern Recognit.
  ({CVPR}), 2020, pp. 6727--6736.

\bibitem{Yin2021}
Y.~Yin, D.~Xu, X.~Wang, L.~Zhang, Agunet: Annotation-guided u-net for fast
  one-shot video object segmentation, Pattern Recognit. 110 (2021) 107580.

\bibitem{Xu2020}
Y.~Xu, Z.~Wang, Z.~Li, et~al., Siamfc++: Towards robust and accurate visual
  tracking with target estimation guidelines, in: {AAAI} Conf. Artif. Intell.
  ({AAAI}), 2020, pp. 12549--12556.

\bibitem{Danelljan2019}
M.~Danelljan, G.~Bhat, F.~S. Khan, M.~Felsberg, {ATOM:} accurate tracking by
  overlap maximization, in: {IEEE} Conf. Comput. Vis. Pattern Recognit.
  ({CVPR}), 2019, pp. 4660--4669.

\bibitem{Bhat2019}
G.~Bhat, M.~Danelljan, L.~V. Gool, R.~Timofte, Learning discriminative model
  prediction for tracking, in: {IEEE} Int. Conf. Comput. Vis. ({ICCV}), 2019,
  pp. 6181--6190.

\bibitem{He2016}
K.~He, X.~Zhang, S.~Ren, J.~Sun, Deep residual learning for image recognition,
  in: {IEEE} Conf. Comput. Vis. Pattern Recognit. ({CVPR}), 2016, pp. 770--778.

\bibitem{He2017}
K.~He, G.~Gkioxari, P.~Doll{\'{a}}r, R.~B. Girshick, Mask {R-CNN}, in: {IEEE}
  Int. Conf. Comput. Vis. ({ICCV}), 2017, pp. 2980--2988.

\bibitem{Lin2017}
T.~Lin, P.~Doll{\'{a}}r, R.~B. Girshick, et~al., Feature pyramid networks for
  object detection, in: {IEEE} Conf. Comput. Vis. Pattern Recognit. ({CVPR}),
  2017, pp. 936--944.

\bibitem{Zhu2018a}
Z.~Zhu, Q.~Wang, B.~Li, et~al., Distractor-aware siamese networks for visual
  object tracking, in: European Conf. Comput. Vis. ({ECCV}), 2018, pp.
  103--119.

\bibitem{Vaswani2017}
A.~Vaswani, N.~Shazeer, N.~Parmar, et~al., Attention is all you need, in: Adv.
  Neural Inf. Process. Syst. ({NIPS}), 2017, pp. 5998--6008.

\bibitem{Kingma2015}
D.~P. Kingma, J.~Ba, Adam: {A} method for stochastic optimization, in: Int.
  Conf. Learn. Repr. ({ICLR}), 2015, pp. 1--15.

\bibitem{Lin2014}
T.~Lin, M.~Maire, S.~J. Belongie, et~al., Microsoft {COCO:} common objects in
  context, in: European Conf. Comput. Vis. ({ECCV}) Workshops, 2014, pp.
  740--755.

\bibitem{Russakovsky2015}
O.~Russakovsky, J.~Deng, H.~Su, et~al., {ImageNet} large scale visual
  recognition challenge, Int. J. Comput. Vision 115~(3) (2015) 211--252.

\bibitem{Real2017}
E.~Real, J.~Shlens, S.~Mazzocchi, et~al., Youtube-boundingboxes: {A} large
  high-precision human-annotated data set for object detection in video, in:
  {IEEE} Conf. Comput. Vis. Pattern Recognit. ({CVPR}), 2017, pp. 7464--7473.

\bibitem{Huang2019}
L.~Huang, X.~Zhao, K.~Huang, Got-10k: {A} large high-diversity benchmark for
  generic object tracking in the wild, {IEEE} Trans. Pattern Anal. Mach.
  Intell. (2019) 1--1.

\bibitem{Milan2016a}
A.~Milan, L.~Leal{-}Taix{\'{e}}, I.~D. Reid, et~al., {MOT}16: {A} benchmark for
  multi-object tracking, CoRR abs/1603.00831.

\bibitem{Yu2020}
H.~Yu, G.~Li, W.~Zhang, et~al., The unmanned aerial vehicle benchmark: Object
  detection, tracking and baseline, Int. J. Comput. Vis. 128~(5) (2020)
  1141--1159.

\bibitem{Zhu2020}
P.~Zhu, L.~Wen, D.~Du, et~al., Vision meets drones: Past, present and future,
  CoRR abs/2001.06303.

\bibitem{Fabbri2018}
M.~Fabbri, F.~Lanzi, S.~Calderara, et~al., Learning to detect and track visible
  and occluded body joints in a virtual world, in: European Conf. Comput. Vis.
  ({ECCV}), 2018, pp. 450--466.

\bibitem{Chen2022}
Z.~Chen, B.~Zhong, G.~Li, et~al., Siamban: Target-aware tracking with siamese
  box adaptive network, {IEEE} Trans. Pattern Anal. Mach. Intell. (2022) 1--17.

\bibitem{Fu2021}
Z.~Fu, Q.~Liu, Z.~Fu, Y.~Wang, Stmtrack: Template-free visual tracking with
  space-time memory networks, in: {IEEE} Conf. Comput. Vis. Pattern Recognit.
  ({CVPR}), 2021, pp. 13774--13783.

\bibitem{Xu2020c}
Y.~Xu, A.~Osep, Y.~Ban, et~al., How to train your deep multi-object tracker,
  in: {IEEE} Conf. Comput. Vis. Pattern Recognit. ({CVPR}), 2020, pp.
  6786--6795.

\bibitem{Shuai2021}
B.~Shuai, A.~G. Berneshawi, X.~Li, et~al., Siammot: Siamese multi-object
  tracking, in: {IEEE} Conf. Comput. Vis. Pattern Recognit. ({CVPR}), 2021, pp.
  12372--12382.

\end{thebibliography}
